\documentclass[letterpaper, twocolumn]{article} 
\usepackage{times}  
\usepackage{helvet}  
\usepackage{courier}  
\usepackage[hyphens]{url}  
\usepackage{graphicx} 
\usepackage[square, sort, comma, numbers]{natbib}  
\usepackage{caption} 
\usepackage{geometry}
\usepackage{algorithm}
\usepackage{algorithmic}
\usepackage{booktabs} 
\usepackage{multirow} 
\usepackage{arydshln}
\usepackage{amssymb}
\usepackage{xspace}
\usepackage{amsmath}
\usepackage{paralist}
\usepackage{mathtools}
\usepackage{hyperref}
\usepackage{nameref}
\usepackage{url}
\usepackage{float}
\usepackage{multicol}
\usepackage{xcolor}
\usepackage{balance}

\newcommand{\aser}[0]{\textup{ASER}\xspace}
\newcommand{\mnv}[0]{\textup{MINERVA}\xspace}
\newcommand{\mhop}[0]{\textup{Multi-Hop}\xspace}
\newcommand{\nelp}[0]{\textup{Neural-LP}\xspace}
\newcommand{\drum}[0]{\textup{DRUM}\xspace}
\newcommand{\anyb}[0]{\textup{AnyBURL}\xspace}
\newcommand{\ptranse}[0]{\textup{PTransE}\xspace}
\newcommand{\rpje}[0]{\textup{RPJE}\xspace}
\newcommand{\transd}[0]{\textup{TransD}\xspace}
\newcommand{\anal}[0]{\textup{ANALOGY}\xspace}
\newcommand{\quate}[0]{\textup{QuatE}\xspace}
\newcommand{\bique}[0]{\textup{BiQUE}\xspace}

\newcommand{\ours}[0]{\textup{BiVE}\xspace}
\newcommand{\oursq}[0]{\textup{BiVE-Q}\xspace}
\newcommand{\oursb}[0]{\textup{BiVE-B}\xspace}

\newcommand{\fb}[0]{\textsl{Freebase}\xspace}
\newcommand{\fbk}[0]{\textsl{FB15K237}\xspace}
\newcommand{\fbko}[0]{\textsl{FB15K}\xspace}
\newcommand{\fbh}[0]{\textsl{FBH}\xspace}
\newcommand{\fbhe}[0]{\textsl{FBHE}\xspace}
\newcommand{\dbp}[0]{\textsl{DBpedia}\xspace}
\newcommand{\dbk}[0]{\textsl{DB15K}\xspace}
\newcommand{\dbhe}[0]{\textsl{DBHE}\xspace}

\newtheorem{mydef}{Definition}

\newcommand{\set}[1]{\mathcal{#1}}
\providecommand{\sE}{\ensuremath{\set{E}}}
\providecommand{\sR}{\ensuremath{\set{R}}}
\providecommand{\sV}{\ensuremath{\set{V}}}
\providecommand{\sH}{\ensuremath{\set{H}}}
\providecommand{\sS}{\ensuremath{\set{S}}}

\providecommand{\sW}{\ensuremath{\set{W}}}

\renewcommand{\vec}[1]{{\bf{#1}}}
\providecommand{\vh}{\ensuremath{\vec{h}}}
\providecommand{\vr}{\ensuremath{\vec{r}}}
\providecommand{\vt}{\ensuremath{\vec{t}}}
\providecommand{\vx}{\ensuremath{\vec{x}}}
\providecommand{\vT}{\ensuremath{\vec{T}}}
\providecommand{\vX}{\ensuremath{\vec{X}}}

\providecommand{\rhat}{\ensuremath{\widehat{r}}}
\providecommand{\Rhat}{\ensuremath{\widehat{\set{R}}}}
\providecommand{\Ehat}{\ensuremath{\widehat{\set{E}}}}
\providecommand{\Ghat}{\ensuremath{\widehat{G}}}

\providecommand{\vrhat}{\ensuremath{\vec{\rhat}}}

\newcommand{\mat}[1]{\boldsymbol{#1}}
\providecommand{\mW}{\ensuremath{\mat{W}}}

\title{\textbf{\Large{Learning Representations of Bi-level Knowledge Graphs for\\Reasoning beyond Link Prediction}\footnote{Published in the Proceedings of the 37th AAAI Conference on Artificial Intelligence (AAAI 2023).}}}
\author{
    \textbf{Chanyoung Chung, Joyce Jiyoung Whang}\footnote{Corresponding author.}
}
\date{
	School of Computing, KAIST\\
	\{chanyoung.chung, jjwhang\}@kaist.ac.kr
}

\renewenvironment{abstract}
 {\small
  \begin{center}
  \bfseries \abstractname\vspace{-.5em}\vspace{0pt}
  \end{center}
  \list{}{%
    \setlength{\leftmargin}{5mm}
    \setlength{\rightmargin}{\leftmargin}%
  }%
  \item\relax}
 {\endlist}

\usepackage{titlesec}
\titleformat{\section}[block]{\large\bfseries\filcenter}{}{1em}{}
\titleformat{\subsection}[hang]{\bfseries}{}{1em}{}
\begin{document}
\maketitle

\begin{abstract}
Knowledge graphs represent known facts using triplets. While existing knowledge graph embedding methods only consider the connections between entities, we propose considering the relationships between triplets. For example, let us consider two triplets $T_1$ and $T_2$ where $T_1$ is (Academy\_Awards, Nominates, Avatar) and $T_2$ is (Avatar, Wins, Academy\_Awards). Given these two base-level triplets, we see that $T_1$ is a prerequisite for~$T_2$. In this paper, we define a higher-level triplet to represent a relationship between triplets, e.g., $\langle T_1$, PrerequisiteFor, $T_2\rangle$ where PrerequisiteFor is a higher-level relation. We define a bi-level knowledge graph that consists of the base-level and the higher-level triplets. We also propose a data augmentation strategy based on the random walks on the bi-level knowledge graph to augment plausible triplets. Our model called \ours learns embeddings by taking into account the structures of the base-level and the higher-level triplets, with additional consideration of the augmented triplets. We propose two new tasks: triplet prediction and conditional link prediction. Given a triplet $T_1$ and a higher-level relation, the triplet prediction predicts a triplet that is likely to be connected to $T_1$ by the higher-level relation, e.g., $\langle T_1$, PrerequisiteFor, ?$\rangle$. The conditional link prediction predicts a missing entity in a triplet conditioned on another triplet, e.g., $\langle T_1$, PrerequisiteFor, (Avatar, Wins, ?)$\rangle$. Experimental results show that \ours significantly outperforms all other methods in the two new tasks and the typical base-level link prediction in real-world bi-level knowledge graphs.
\end{abstract}
\vspace{-0.2cm}
\section{Introduction}
\vspace{-0.2cm}
\label{sec:intro}
A knowledge graph represents the relationships between entities using triplets consisting of a head entity, a relation, and a tail entity. Knowledge graph embedding aims to represent the entities and relations as a set of embedding vectors that can be utilized in many modern AI applications~\cite{kgss,robo}. Most existing knowledge graph embedding methods generate the embedding vectors by focusing solely on how the entities are connected by the relations~\cite{kgs,meta,atth}. Even though some methods predict missing connections between the entities by rule mining~\cite{anyb,drum} or rule-and-path-based learning~\cite{rpje}, these existing approaches only enable expanding the entity-level connections.

\begin{figure}[t]
\centering
\includegraphics[width = \columnwidth]{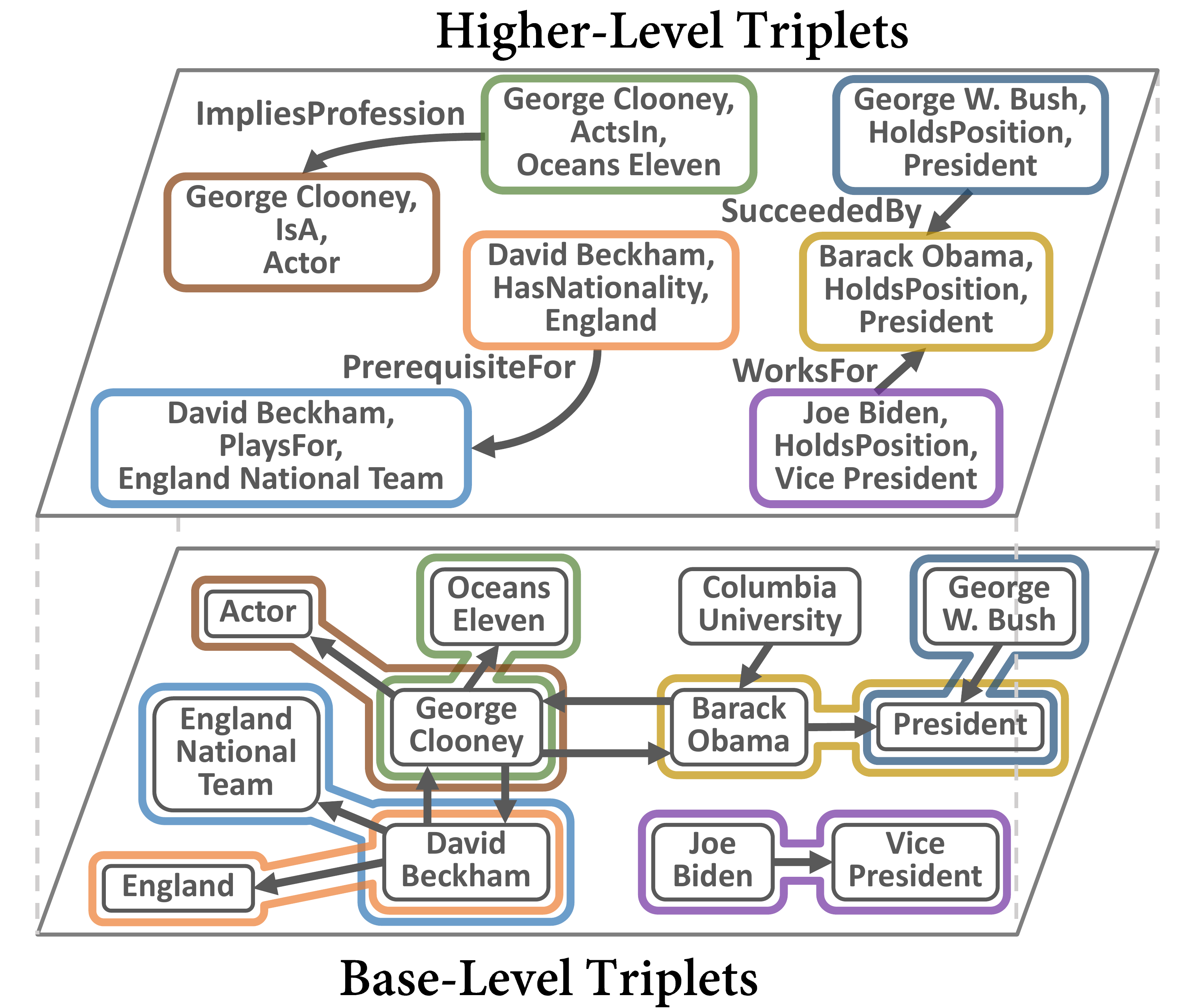}
\caption{Example of a bi-level knowledge graph consisting of base-level and higher-level triplets in the \fbhe dataset. The relation labels are omitted in the base-level triplets.}
\label{fig:bikg}
\vspace{-0.5cm}
\end{figure}

Each triplet in a knowledge graph can have a relationship with another triplet. For example, let us consider two base-level triplets $T_1$ and $T_2$ where $T_1$ is (Joe\_Biden, HoldsPosition, Vice\_President) and $T_2$ is (Barack\_Obama, HoldsPosition, President). To represent the fact that Joe Biden was a vice president when Barack Obama was a president, we define a higher-level triplet $\langle T_1$, WorksFor, $T_2\rangle$ where WorksFor is a higher-level relation. In this paper, we define a bi-level knowledge graph that includes both the base-level and the higher-level triplets, where the base-level triplets correspond to the original triplets representing the relationships between entities, while the higher-level triplets represent the relationships between the base-level triplets using the higher-level relations. Based on well-known knowledge graphs, \fbk~\cite{fb} and \dbk~\cite{kblrn}, we create three real-world bi-level knowledge graphs named \fbh, \fbhe, and \dbhe. Figure~\ref{fig:bikg} shows a subgraph of a bi-level knowledge graph in \fbhe where the base-level triplets correspond to the original triplets in \fbk and the higher-level triplets are manually created by defining the higher-level relationships between the base-level triplets.

We propose incorporating the base-level and the higher-level triplets into knowledge graph embedding. Using the bi-level knowledge graphs, we also propose a data augmentation strategy that augments triplets by identifying plausible relation sequences based on random walks. We develop a new knowledge graph embedding method called \ours (embedding of {\bf\textsf{Bi}}-le{\bf\textsf{V}}el knowledg{\bf\textsf{E}} graphs) that computes embedding vectors by reflecting the structures of the base-level and the higher-level triplets simultaneously, where the augmented triplets are further incorporated. Using the bi-level knowledge graphs, we propose two new tasks: triplet prediction and conditional link prediction. The triplet prediction predicts a triplet that is likely to be connected to a given triplet using a higher-level relation, e.g., $\langle T_1$, WorksFor, ?$\rangle$, whereas the conditional link prediction predicts a missing entity in a triplet where another triplet is provided as a condition, e.g., $\langle T_1$, WorksFor, (?, HoldsPosition, President)$\rangle$. Experimental results show that \ours significantly outperforms other state-of-the-art knowledge graph completion methods in real-world datasets.\footnote{\url{https://github.com/bdi-lab/BiVE}} 

Our contributions can be summarized as follows:
\begin{compactitem}
\item To the best of our knowledge, our work is the first work that introduces the higher-level relationships between triplets in knowledge graphs; we define bi-level knowledge graphs and create three real-world datasets.
\item We propose an efficient data augmentation strategy using random walks on a bi-level knowledge graph.
\item We develop \ours to learn embeddings by effectively incorporating the base-level triplets, the higher-level triplets, and the augmented triplets.
\item We propose two new tasks, triplet prediction and conditional link prediction, which have never been studied. 
\item \ours significantly outperforms 12 different state-of-the-art knowledge graph completion methods.
\end{compactitem}
\vspace{-0.2cm}
\section{Related Work}
\vspace{-0.2cm}
Some knowledge graph completion methods use multi-hop paths between distant entities~\cite{rpje,pte,ptd,mhop,mnv} and rule-based or logic-based methods identify frequently observed patterns~\cite{anyb,inject,kale,nelp,drum,lenn}. The main difference between these methods and \ours is that the existing methods only consider the relationships between entities, whereas \ours considers not only the relationships between entities but also the relationships between triplets. Also, the way of expressing the relationships between entities or triplets in \ours is not restricted to the first-order-logic-like expression. For example, the rule-based methods consider the relationships between connected entities, e.g., $\forall x,y,z: (x,r_1,y)\wedge (y,r_2,z) \Rightarrow (x,r_3,z)$ where there should exist a path connecting $x$, $y$, and $z$ in the knowledge graph. On the other hand, \ours represents relationships like $(x, r_1, y) \xRightarrow{\rhat} (p, r_2, q)$ where $x$, $y$ and $p$, $q$ are not necessarily connected by the base-level triplets, and also $r_1$, $r_2$, and $\xRightarrow{\rhat}$ can be any relation not restricted to the first-order-logic-like relation. 

Even though there have been many attempts to discover meaningful patterns in a knowledge graph and utilize them to complete missing links~\cite{pra}, such attempts have rarely been studied in the context of data augmentation. Recently, rule-based data augmentation for knowledge graph embedding has been proposed~\cite{daug}\footnote{We could not include this method as a baseline in our experiments because the authors of~\cite{daug} could not provide their source codes due to some confidentiality restrictions.}. While this method uses logical rules using the base-level triplets, our data augmentation employs random walks on a bi-level knowledge graph. 

To exploit enriched information about triplets, some knowledge graph embedding methods utilize attributes of entities or ontological concepts~\cite{joie}. TransEA~\cite{transea} considers numeric attributes of entities, and LiteralE~\cite{literale} considers information from literals. HINGE~\cite{hinge} has been proposed to represent hyper-relational facts where a triplet has additional key-value pairs to present extra information about each triplet. Even though these methods consider enriched information about triplets, they do not consider the relationships between triplets.

In information retrieval, a neural fact contextualization method has been proposed to rank a set of candidate facts for a given triplet~\cite{cvt}. Also, a way of representing a triplet in an embedding space is studied by considering the concept of a line graph~\cite{ttv}. Recently, ATOMIC~\cite{atomic} has been proposed to provide commonsense knowledge for if-then reasoning, whereas \aser~\cite{aser} has been proposed to construct an eventuality knowledge graph. Although these methods consider triplet-level operations, the goal of their methods is different from ours and none of these considers the bi-level knowledge graphs.
\vspace{-0.2cm}
\section{Bi-Level Knowledge Graphs}
\vspace{-0.2cm}
Let us represent a knowledge graph as $G=(\sV,\sR,\sE)$ where $\sV$ is a set of entities, $\sR$ is a set of relations, and $\sE=\{(h,r,t) : h\in\sV, r\in\sR, t\in\sV\}$ is a set of triplets. Let us call $G$ a base-level knowledge graph and call $(h,r,t)\in\sE$ a base-level triplet. We formally define the higher-level triplets as follows.

\begin{mydef}
[Higher-Level Triplets] Given a base-level knowledge graph $G=(\sV,\sR,\sE)$, a set of higher-level triplets is defined by $\sH=\{\langle T_i, \rhat, T_j\rangle:T_i\in\sE, \rhat\in\Rhat, T_j\in\sE \}$ where $\sE$ is a set of base-level triplets and $\Rhat$ is a set of higher-level relations connecting the base-level triplets.
\end{mydef}

We define a bi-level knowledge graph as follows. 

\begin{mydef}
[Bi-Level Knowledge Graph] Given a base-level knowledge graph $G=(\sV,\sR,\sE)$, a set of higher-level relations $\Rhat$, and a set of higher-level triplets $\sH$, a bi-level knowledge graph is defined as $\Ghat=(\sV, \sR, \sE, \Rhat, \sH)$. 
\end{mydef}

To define a bi-level knowledge graph, we add the higher-level triplets $\sH$ to the base-level knowledge graph $G$ by introducing the higher-level relations $\Rhat$. We create real-world bi-level knowledge graphs \fbh and \fbhe based on \fbk from \fb~\cite{fbl} and \dbhe based on \dbk from \dbp~\cite{dbp}. Table~\ref{tb:hlt} shows some examples of the higher-level relations and triplets. \fbh contains the higher-level relations that can be inferred inside the base-level knowledge graph, e.g., PrerequisiteFor and ImpliesProfession, whereas \fbhe and \dbhe contain some externally-sourced knowledge, e.g., WorksFor and NextAlmaMater. For example, we crawl Wikipedia articles to find information about the (vice)presidents of the United States and the alma mater information of politicians. As a result, \fbh contains six different higher-level relations, \fbhe has ten higher-level relations, and \dbhe has eight higher-level relations. Note that the base-level knowledge graphs of \fbh and \fbhe are \fbk. \fbhe extends \fbh by including the externally-sourced higher-level relationships. The authors of this paper manually defined the higher-level relations and added the higher-level triplets to \fbk and \dbk, which took six weeks. 


\begin{table}[t]
\scriptsize
\centering
\setlength{\tabcolsep}{0.3em}
\begin{tabular}{ccl}
\toprule
& $\rhat$ & \multicolumn{1}{c}{$\langle T_i,\rhat,T_j \rangle$} \\
\midrule
\multirow{8}{*}{\fbhe} & \multirow{2}{*}{PrerequisiteFor} & $T_i$: (BAFTA\_Award, Nominates, The\_King's\_Speech) \\
 & & $T_j$: (The\_King's\_Speech, Wins, BAFTA\_Award) \\
\cdashline{2-3}
 & \multirow{2}{*}{ImpliesProfession} & $T_i$: (Liam\_Neeson, ActsIn, Love\_Actually) \\
 & & $T_j$: (Liam\_Neeson, IsA, Actor) \\
\cdashline{2-3}
 & \multirow{2}{*}{WorksFor} & $T_i$: (Joe\_Biden, HoldsPosition, Vice\_President) \\
 & & $T_j$: (Barack\_Obama, HoldsPosition, President) \\
\cdashline{2-3}
 & \multirow{2}{*}{SucceededBy} & $T_i$: (George\_W.\_Bush, HoldsPosition, President) \\
 & & $T_j$: (Barack\_Obama, HoldsPosition, President) \\
\cdashline{1-3}
\multirow{4}{*}{\dbhe} & \multirow{2}{*}{ImpliesTimeZone} & $T_i$: (Czech\_Republic, TimeZone, Central\_European) \\
 & & $T_j$: (Prague, TimeZone, Central\_European) \\
\cdashline{2-3}
 & \multirow{2}{*}{NextAlmaMater} & $T_i$: (Gerald\_Ford, StudiesIn, University\_of\_Michigan) \\
 & & $T_j$: (Gerald\_Ford, StudiesIn, Yale\_University) \\
\bottomrule
\end{tabular}
\vspace{-0.2cm}
\caption{Examples of Higher-Level Relations and Triplets.}
\label{tb:hlt}
\end{table}

Using a bi-level knowledge graph, we define the triplet prediction problem as follows.

\begin{mydef}
[Triplet Prediction] Given a bi-level knowledge graph $\Ghat=(\sV, \sR, \sE, \Rhat, \sH)$ where $\sH=\{\langle T_i, \rhat, T_j\rangle:T_i\in\sE, \rhat\in\Rhat, T_j\in\sE \}$, the triplet prediction problem is defined as $\langle T_i, \rhat, ?\rangle$ or $\langle ?, \rhat, T_j\rangle$ where the goal is to predict the missing base-level triplet.
\end{mydef}

Also, we define the conditional link prediction as follows.

\begin{mydef}
[Conditional Link Prediction] Given a bi-level knowledge graph $\Ghat=(\sV, \sR, \sE, \Rhat, \sH)$ where $\sH=\{\langle T_i, \rhat, T_j\rangle:T_i\in\sE, \rhat\in\Rhat, T_j\in\sE \}$, let $T_i\coloneqq(h_i,r_i,t_i)$ and $T_j\coloneqq(h_j,r_j,t_j)$. The conditional link prediction problem is to predict a missing entity in a base-level triplet conditioned on another base-level triplet. Specifically, the problem is defined as $\langle T_i, \rhat, (h_j,r_j,?) \rangle$ or $\langle T_i, \rhat, (?,r_j,t_j) \rangle$ or $\langle (h_i,r_i,?), \rhat, T_j \rangle$ or $\langle (?,r_i,t_i), \rhat, T_j \rangle$.
\end{mydef}
\vspace{-0.2cm}
\section{Data Augmentation by Random Walks on a Bi-Level Knowledge Graph}
\vspace{-0.2cm}
Consider a bi-level knowledge graph in the training set $\Ghat_{\text{train}}=(\sV, \sR, \sE_{\text{train}}, \Rhat, \sH_{\text{train}})$ where $\sE_{\text{train}}$ and $\sH_{\text{train}}$ are the base-level and the higher-level triplets in the training set respectively. We add reverse relations to $\sR$ and add reversed triplets to $\sE_{\text{train}}$, i.e., for every $r\in\sR$, we add $r^{-1}$ that has the reverse direction of $r$ and for every $(h,r,t)\in\sE_{\text{train}}$, we add $(t,r^{-1},h)$ to $\sE_{\text{train}}$. Similarly, for every $\rhat\in\Rhat$, we add $\rhat^{-1}$ and add the reversed higher-level triplets to $\sH_{\text{train}}$. All these reverse relations and reversed triplets are added only for data augmentation.

From an entity $h$, we randomly visit one of its neighbors by following a base-level or a higher-level triplet. To search for diverse patterns, we do not allow going back to an entity that has already been visited. Let us define a random walk path to be the sequence of visited entities, visited relations, and visited higher-level relations. Consider two base-level triplets $T_i=(h_i,r_i,t_i)$ and $T_j=(h_j,r_j,t_j)$ and a higher-level triplet $\langle T_i, \rhat, T_j \rangle$. From any entity in $T_i$, we can go to any entity in $T_j$ and vice versa by following $r_i$, $\rhat$, and $r_j$ or their reverse relations. For example, one possible random walk path is $(h_i,r_i,\rhat,r_j,t_j)$. Another possible random walk path is $(t_j,r_j^{-1},\rhat^{-1},r_i,t_i)$. Assume that we have a base-level triplet $T_0=(h_0,r_0,h_i)$. Starting from $h_0$, we can make a longer path, e.g., $(h_0,r_0,h_i,r_i,\rhat,r_j,t_j)$. We define the length of a random walk path to be the number of entities in the sequence except the starting entity. 

Given the maximum length of a random walk path $L$, we repeat the random walks by varying the length $l=2,\cdots{},L$ and repeat the random walks $n$ times for every $l$. In our experiments, we set $L$=3 and $n$=50,000,000. Let $w$ denote the sequence of a random walk path of all possible lengths, where we randomly select a starting entity for every $w$. If there are multiple identical random walk paths, we remove the duplicates to prevent unexpected bias. Let $p_k$ be the $k$-th unique sequence of relations and higher-level relations extracted from $w$, i.e., we make $p_k$ by removing all entities from $w$, e.g., if $w=(h_0,r_0,h_i,r_i,\rhat,r_j,t_j)$ then $p_k=(r_0,r_i,\rhat,r_j)$. We call $p_k$ the relation sequence. Since $p_k$ only traces the relations, different random walk paths can be mapped into the same $p_k$. Using $p_k$, we rewrite a random walk path $w=(h,\cdots{},t)$ to $w=(h,p_k,t)$ where the relation sequence of the original path $w$ is mapped into $p_k$, $h$ is the starting entity and $t$ is the last entity. Let $\sW$ denote the multiset of all random walk paths of all possible lengths. We define the confidence score of $(p_k,r)$ as 

\begin{displaymath}
\label{eq:conf}
c(p_k,r) \coloneqq \dfrac{|\{ (h,r,t) : (h,p_k,t)\in\sW, (h,r,t)\in\sE_{\text{train}} \}|}{|\{(h,p_k,t) :  (h,p_k,t)\in\sW\}|}.
\end{displaymath}

\begin{figure}[t]
\centering
\includegraphics[width = 0.8\columnwidth]{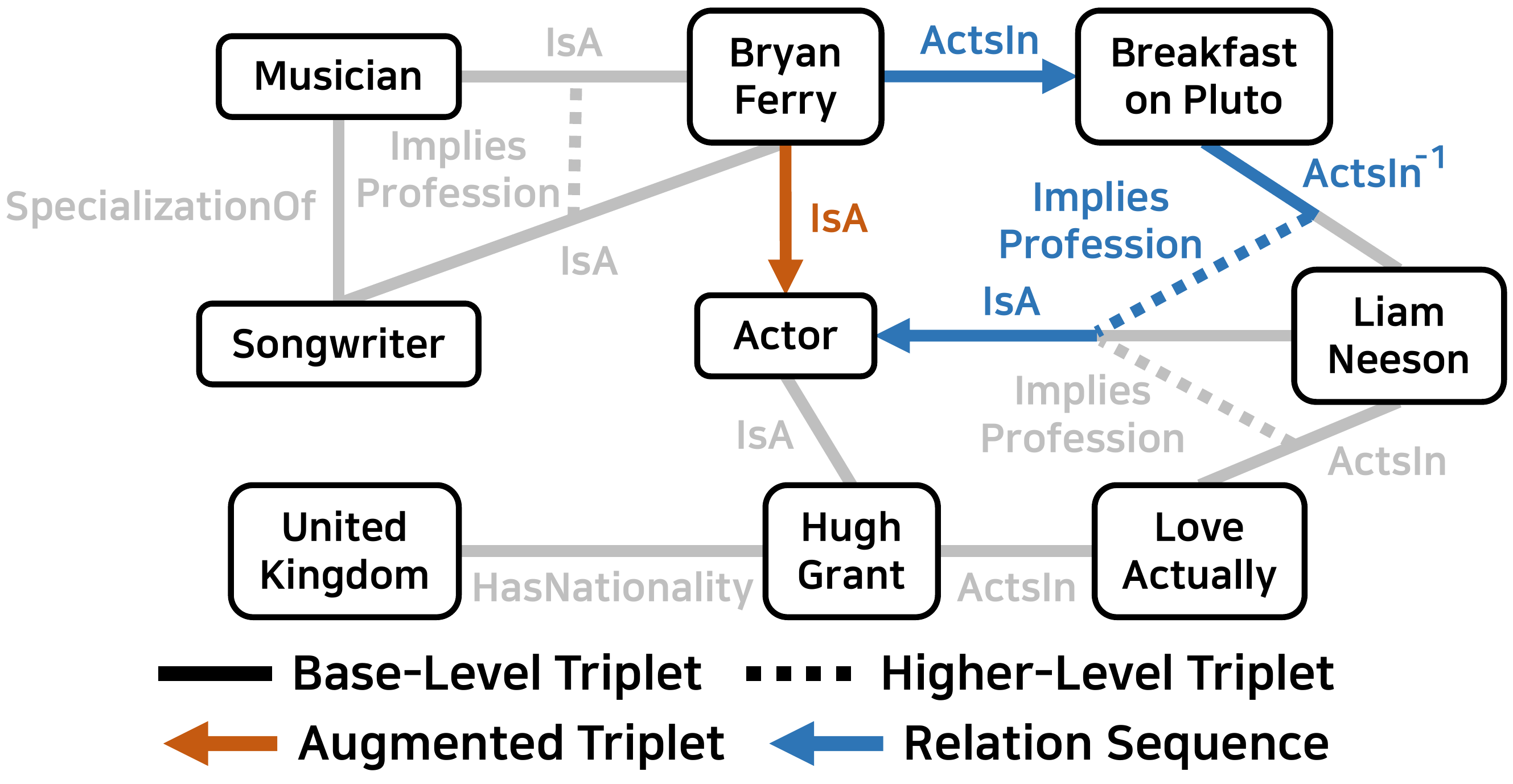}
\vspace{-0.2cm}
\caption{Random walk path in a bi-level knowledge graph and an augmented triplet in \fbh. We add missing triplets whose confidence scores are greater than a certain threshold.}
\label{fig:path}
\vspace{-0.3cm}
\end{figure}

We select the pairs of $(p_k,r)$ that satisfies $c(p_k,r)\geq \tau$ where we set $\tau=0.7$. Let $\sS_{kr}\coloneqq \{ (h,r,t) : (h,p_k,t)\in\sW, c(p_k,r)\geq \tau, (h,r,t)\notin\sE_{\text{train}}\}$ where $\sS_{kr}$ indicates a set of missing triplets $(h,r,t)$ even though $c(p_k,r)\geq \tau$. Then, let $\sS \coloneqq \cup_k \cup_r \sS_{kr}$ where $\sS$ is a set of augmented triplets. We add the triplets in $\sS$ to a bi-level knowledge graph to augment triplets that are likely to be present. Figure~\ref{fig:path} shows an example of a random walk path of length $2$ and an augmented triplet in \fbh, where the walk starts from Bryan\_Ferry. Let $p_1= $ (ActsIn, ActsIn$^{-1}$, ImpliesProfession, IsA). Since the confidence score of ($p_1$, IsA) is $0.99$, we add a triplet (Bryan\_Ferry, IsA, Actor) which was missing in the original training set.  

\vspace{-0.2cm}
\section{Embedding of Bi-Level Knowledge Graphs}
\vspace{-0.2cm}
A knowledge graph embedding method defines a scoring function $f(\vh,\vr,\vt)$ of a triplet $(h,r,t)$, where $\vh$, $\vr$, and $\vt$ are embedding vectors of $h$, $r$, and $t$ respectively; a higher score indicates a more plausible triplet. In \ours, the loss incurred by the base-level triplets, $L_{\text{base}}$, is defined as follows:

{\scriptsize
\begin{displaymath}
L_{\text{base}}\coloneqq\sum_{(h,r,t)\in\sE_{\text{train}}}g(-f(\vh,\vr,\vt))+\sum_{(h',r',t')\in\sE_{\text{\text{train}}}'}g(f(\vh',\vr',\vt'))
\end{displaymath}}where $g(x)=\text{log}(1+\text{exp}(x))$ and $\sE_{\text{\text{train}}}'$ is a set of corrupted triplets. We can use any knowledge graph embedding scoring function for $f(\cdot)$. We implement \ours with two different scoring functions for $f(\cdot)$: \quate~\cite{quate} for \oursq and \bique~\cite{bique} for \oursb.

Given $T_i=(h_i,r_i,t_i)$, let $\vT_i$ denote an embedding vector of $T_i$ where the dimension is $\hat{d}$. We define $\vT_i\coloneqq\mW[\vh_i;\vr_i;\vt_i]$ where $\vh_i$, $\vr_i$, and $\vt_i$ denote the embedding vectors of $h_i$, $r_i$, and $t_i$ respectively, the dimension of each of these embedding vectors is $d$, and $\mW$ is a projection matrix of size $\hat{d}\times 3d$ which projects the vertically concatenated vector to the $\hat{d}$-dimensional space. Similarly, $\vT_j=\mW[\vh_j;\vr_j;\vt_j]$ where $T_j=(h_j,r_j,t_j)$. We define the loss incurred by the higher-level triplets, $L_{\text{high}}$, as follows:

{\scriptsize
\begin{displaymath}
L_{\text{high}}\coloneqq\sum_{\langle T_i,\rhat,T_j\rangle}g(-f(\vT_i,\vrhat,\vT_j))+\sum_{\langle{T_i}',\rhat',{T_j}'\rangle}g(f(\vT_i',\vrhat',\vT_j'))
\end{displaymath}}where $\langle T_i,\rhat,T_j\rangle\in\sH_{\text{train}}$, $\langle{T_i}',\rhat',{T_j}'\rangle\in\sH_{\text{train}}'$, $\vrhat$ is the embedding vector of $\rhat$, the dimension of $\vrhat$ is $\hat{d}$, and $\langle{T_i}',\rhat',{T_j}'\rangle$ is a corrupted higher-level triplet made by randomly replacing $T_i$ or $T_j$ with one of the triplets in $\sE_{\text{train}}$. 

We define the loss of the augmented triplets, $L_{\text{aug}}$, as

{\scriptsize
\begin{displaymath}
L_{\text{aug}}\coloneqq\sum_{(h,r,t)\in\sS}g(-f(\vh,\vr,\vt))+\sum_{(h',r',t')\in\sS'}g(f(\vh',\vr',\vt'))
\end{displaymath}}where $\sS$ is the set of the augmented triplets and $\sS'$ is the set of corrupted triplets.

Finally, our loss function of \ours is defined by
\begin{displaymath}
L_{\text{\ours}} \coloneqq L_{\text{base}} + \lambda_1 \cdot L_{\text{high}} + \lambda_2 \cdot L_{\text{aug}}
\end{displaymath} where $\lambda_1$ is a hyperparameter indicating the importance of the higher-level triplets and $\lambda_2$ indicates the importance of the augmented triplets. By optimizing $L_{\text{\ours}}$, \ours learns embeddings by considering the structures of the base-level triplets, the higher-level triplets, and the augmented triplets.

Let us describe the scoring functions of \ours for triplet prediction and conditional link prediction. To solve a triplet prediction problem, $\langle T_i, \rhat, ? \rangle$, we compute $F_{\text{tp}}(X) \coloneqq f(\vT_i,\vrhat,\vX)$ for every base-level triplet $X\in\sE_{\text{train}}$ where $\vX$ is a learned embedding vector of $X$. To solve a conditional link prediction problem, $\langle T_i, \widehat{r}, (h_j,r_j,?) \rangle$, we compute $F_{\text{clp}}(x) \coloneqq f(\vh_j,\vr_j,\vx) + \lambda_1 \cdot f(\vT_i,\vrhat,\mW[\vh_j;\vr_j;\vx])$ for every $x\in\sV$ where $\vx$ is a learned embedding of $x$. 

\begin{table}[t]
\small
\centering
\setlength{\tabcolsep}{0.65em}
\begin{tabular}{ccccccc}
\toprule
 & $|\sV|$ & $|\sR|$ & $|\sE|$ & $|\Rhat|$ & $|\sH|$ & $|\Ehat|$ \\
\midrule
\fbh & 14,541 & 237 & 310,116 & 6 & 27,062 & 33,157 \\
\fbhe & 14,541 & 237 & 310,116 & 10 & 34,941 & 33,719 \\
\dbhe & 12,440 & 87 & 68,296 & 8 & 6,717 & 8,206 \\
\bottomrule
\end{tabular}
\vspace{-0.2cm}
\caption{Statistics of a bi-level knowledge graph $\Ghat=(\sV, \sR, \sE, \Rhat, \sH)$. $|\Ehat|$ is the number of base-level triplets which are involved in the higher-level triplets.}
\label{tb:data}
\vspace{-0.8cm}
\end{table}


\begin{table*}[t]
\small
\centering
\setlength{\tabcolsep}{0.22em}
\begin{tabular}{cccccccccc}
\toprule
 & \multicolumn{3}{c}{\fbh} & \multicolumn{3}{c}{\fbhe} & \multicolumn{3}{c}{\dbhe} \\
 & MR ($\downarrow$) & MRR ($\uparrow$) & Hit@10 ($\uparrow$) & MR ($\downarrow$) & MRR ($\uparrow$) & Hit@10 ($\uparrow$) & MR ($\downarrow$) & MRR ($\uparrow$) & Hit@10 ($\uparrow$) \\
\midrule
\aser & 74541.7{\scriptsize$\pm$0.0} & 0.011{\scriptsize$\pm$0.000} & 0.015{\scriptsize$\pm$0.000} & 57916.0{\scriptsize$\pm$0.0} & 0.050{\scriptsize$\pm$0.000} & 0.070{\scriptsize$\pm$0.000} & 18157.6{\scriptsize$\pm$0.0} & 0.042{\scriptsize$\pm$0.000} & 0.075{\scriptsize$\pm$0.000} \\
\mnv & 109055.1{\scriptsize$\pm$98.5} & 0.093{\scriptsize$\pm$0.002} & 0.113{\scriptsize$\pm$0.002} & 85571.5{\scriptsize$\pm$768.3} & 0.220{\scriptsize$\pm$0.008} & 0.300{\scriptsize$\pm$0.005} & 20764.3{\scriptsize$\pm$72.3} & 0.177{\scriptsize$\pm$0.005} & 0.221{\scriptsize$\pm$0.004} \\
\mhop & 108731.7{\scriptsize$\pm$43.2} & 0.105{\scriptsize$\pm$0.001} & 0.117{\scriptsize$\pm$0.000} & 83643.8{\scriptsize$\pm$33.2} & 0.255{\scriptsize$\pm$0.012} & 0.311{\scriptsize$\pm$0.003} & 20505.8{\scriptsize$\pm$9.3} & 0.191{\scriptsize$\pm$0.001} & 0.230{\scriptsize$\pm$0.002} \\
\nelp & 115016.6{\scriptsize$\pm$0.0} & 0.070{\scriptsize$\pm$0.000} & 0.073{\scriptsize$\pm$0.000} & 90000.4{\scriptsize$\pm$0.0} & 0.238{\scriptsize$\pm$0.000} & 0.274{\scriptsize$\pm$0.000} & 21130.5{\scriptsize$\pm$0.0} & 0.170{\scriptsize$\pm$0.000} & 0.209{\scriptsize$\pm$0.000} \\
\drum & 115016.6{\scriptsize$\pm$0.0} & 0.069{\scriptsize$\pm$0.001} & 0.073{\scriptsize$\pm$0.000} & 90000.3{\scriptsize$\pm$0.0} & 0.261{\scriptsize$\pm$0.000} & 0.274{\scriptsize$\pm$0.000} & 21130.5{\scriptsize$\pm$0.0} & 0.166{\scriptsize$\pm$0.001} & 0.209{\scriptsize$\pm$0.000} \\
\anyb & 108079.6{\scriptsize$\pm$0.0} & 0.096{\scriptsize$\pm$0.000} & 0.108{\scriptsize$\pm$0.000} & 83136.8{\scriptsize$\pm$5.3} & 0.191{\scriptsize$\pm$0.001} & 0.252{\scriptsize$\pm$0.001} & 20530.8{\scriptsize$\pm$0.0} & 0.177{\scriptsize$\pm$0.000} & 0.214{\scriptsize$\pm$0.000} \\
\ptranse & 111024.3{\scriptsize$\pm$855.0} & 0.069{\scriptsize$\pm$0.000} & 0.071{\scriptsize$\pm$0.000} & 86793.2{\scriptsize$\pm$961.0} & 0.249{\scriptsize$\pm$0.001} & 0.274{\scriptsize$\pm$0.000} & 18888.7{\scriptsize$\pm$457.3} & 0.158{\scriptsize$\pm$0.001} & 0.195{\scriptsize$\pm$0.002} \\
\rpje & 113082.0{\scriptsize$\pm$945.2} & 0.070{\scriptsize$\pm$0.000} & 0.072{\scriptsize$\pm$0.000} & 89173.1{\scriptsize$\pm$797.3} & 0.267{\scriptsize$\pm$0.000} & 0.274{\scriptsize$\pm$0.000} & 20290.4{\scriptsize$\pm$417.2} & 0.166{\scriptsize$\pm$0.001} & 0.206{\scriptsize$\pm$0.002} \\
\transd & 74277.3{\scriptsize$\pm$2907.8} & 0.052{\scriptsize$\pm$0.001} & 0.104{\scriptsize$\pm$0.002} & 52159.4{\scriptsize$\pm$758.9} & 0.238{\scriptsize$\pm$0.002} & 0.280{\scriptsize$\pm$0.003} & 16698.1{\scriptsize$\pm$370.2} & 0.116{\scriptsize$\pm$0.004} & 0.189{\scriptsize$\pm$0.009} \\
\anal & 93383.4{\scriptsize$\pm$20576.5} & 0.072{\scriptsize$\pm$0.004} & 0.107{\scriptsize$\pm$0.002} & 60161.5{\scriptsize$\pm$3295.5} & 0.286{\scriptsize$\pm$0.004} & 0.318{\scriptsize$\pm$0.001} & 18880.0{\scriptsize$\pm$1213.8} & 0.150{\scriptsize$\pm$0.005} & 0.211{\scriptsize$\pm$0.005} \\
\quate & 145603.8{\scriptsize$\pm$1114.4} & 0.103{\scriptsize$\pm$0.001} & 0.114{\scriptsize$\pm$0.001} & 94684.4{\scriptsize$\pm$1781.7} & 0.101{\scriptsize$\pm$0.009} & 0.209{\scriptsize$\pm$0.011} & 26485.0{\scriptsize$\pm$491.8} & 0.157{\scriptsize$\pm$0.003} & 0.179{\scriptsize$\pm$0.002} \\
\bique & 81687.5{\scriptsize$\pm$603.2} & 0.104{\scriptsize$\pm$0.000} & 0.115{\scriptsize$\pm$0.000} & 61015.2{\scriptsize$\pm$399.8} & 0.135{\scriptsize$\pm$0.002} & 0.205{\scriptsize$\pm$0.007} & 19079.4{\scriptsize$\pm$389.7} & 0.163{\scriptsize$\pm$0.002} & 0.185{\scriptsize$\pm$0.002} \\
\oursq & \textbf{18.7{\scriptsize$\pm$1.2}} & \textbf{0.748{\scriptsize$\pm$0.007}} & \textbf{0.853{\scriptsize$\pm$0.004}} & \underline{33.1{\scriptsize$\pm$17.4}} & \underline{0.531{\scriptsize$\pm$0.106}} & \underline{0.683{\scriptsize$\pm$0.114}} & \underline{56.6{\scriptsize$\pm$10.2}} & \underline{0.315{\scriptsize$\pm$0.024}} & \underline{0.523{\scriptsize$\pm$0.034}} \\
\oursb & \underline{19.7{\scriptsize$\pm$1.9}} & \underline{0.731{\scriptsize$\pm$0.010}} & \underline{0.837{\scriptsize$\pm$0.006}} & \textbf{27.9{\scriptsize$\pm$2.4}} & \textbf{0.555{\scriptsize$\pm$0.007}} & \textbf{0.718{\scriptsize$\pm$0.007}} & \textbf{4.7{\scriptsize$\pm$0.2}} & \textbf{0.644{\scriptsize$\pm$0.004}} & \textbf{0.914{\scriptsize$\pm$0.005}} \\
\bottomrule
\end{tabular}
\vspace{-0.2cm}
\caption{Results of Triplet Prediction. The best scores are boldfaced and the second best scores are underlined. Our models, \oursq and \oursb, significantly outperform all other baseline methods in terms of all metrics on all datasets.}
\label{tb:tp}
\vspace{-0.2cm}
\end{table*}
\section{Experimental Results\protect\footnote{Results with a new implementation of \ours are provided in Appendix \hyperref[sec:new]{Additional Experimental Results}. Although the original implementation is also correct, the new implementation improves the performance of \ours.}}
\vspace{-0.2cm}
We use three real-world bi-level knowledge graphs presented in Table~\ref{tb:data}, where $|\Ehat|$ is the number of base-level triplets involved in the higher-level triplets. We split $\sE$ and $\sH$ into training, validation, and test sets with a ratio of 8:1:1. We use three standard evaluation metrics: the filtered MR (Mean Rank), MRR (Mean Reciprocal Rank), and Hit@$10$~\cite{kgs}. Higher MRR and Hit@$10$ and a lower MR indicate better results. We repeat experiments ten times for each method and report the mean and the standard deviation. We set $d=200$ and $\hat{d}=200$. We use 12 different baseline methods: \aser~\cite{aser}, \mnv~\cite{mnv}, \mhop~\cite{mhop}, \nelp~\cite{nelp}, \drum~\cite{drum}, \anyb~\cite{anyb}, \ptranse~\cite{pte}, \rpje~\cite{rpje}, \transd~\cite{transd}, \anal~\cite{anal}, \quate~\cite{quate} and \bique~\cite{bique}. For \transd and \anal, we use the implementations in OpenKE~\cite{openke}. More details of datasets and methods are described in {\hyperref[sec:app]{Appendix}}. 



\begin{figure}[t]
\centering
\includegraphics[width = 0.7\columnwidth]{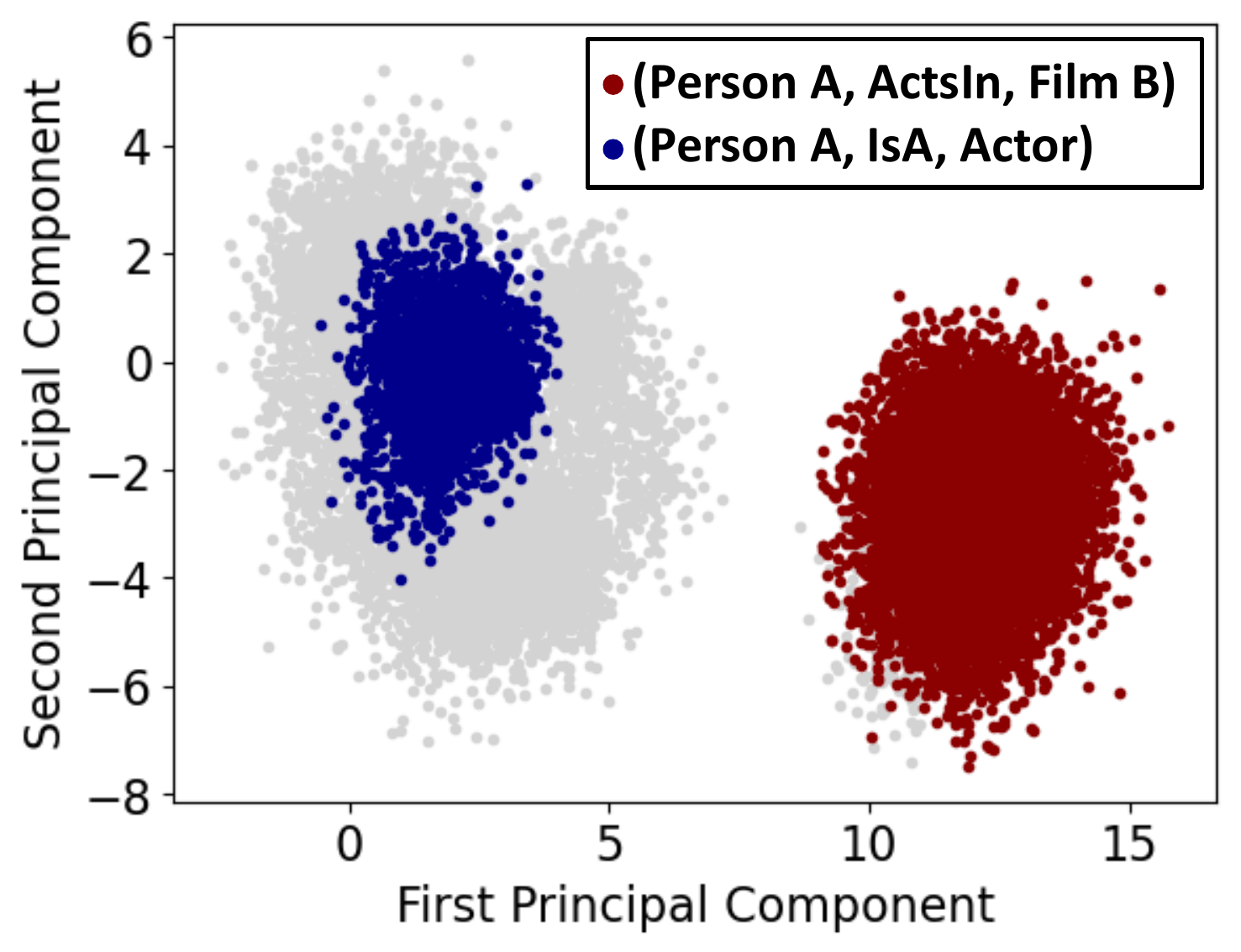}
\vspace{-0.2cm}
\caption{Embedding vectors of base-level triplets in $\langle T_i$, ImpliesProfession, $T_j\rangle$ where $T_i$ is (Person~A, ActsIn, Film~B) and $T_j$ is (Person~A, IsA, Actor) in \fbh. Both $\vT_i$ and $\vT_j$ embedding vectors from \ours are well-clustered.}
\label{fig:tri_emb}
\vspace{-0.5cm}
\end{figure}
\vspace{-0.2cm}
\subsection{Triplet Prediction}
\vspace{-0.2cm}
While \ours solves a triplet prediction problem using the scoring function $F_{\text{tp}}(X)$, none of the baseline methods can deal with the higher-level triplets. To feed the higher-level triplets to the baseline methods, we create a new knowledge graph $G_T$ where a base-level triplet is converted into an entity and a higher-level triplet is converted into a triplet. Let $T_i=(h_i,r_i,t_i)\in\sE_{\text{train}}$ denote a base-level triplet. We define $G_T\coloneqq (\sE_{\text{train}},\Rhat,\sH_{\text{train}})$, where each $T_i$ is considered as an entity. If we train the baseline methods using $G_T$, the triplet prediction task can be considered as a link prediction problem on $G_T$. However, in this case, it is not guaranteed that all $T_i$ involved in the triplets in $\sH_{\text{test}}$ appear in $\sH_{\text{train}}$ because we randomly split $\sH$ into training, validation, and test sets. Therefore, for the baseline methods, the problem becomes an inductive setting instead of a transductive setting. Indeed, among the baseline methods, \nelp and \drum are inductive methods and we include these methods because they can conduct inductive inference. We assume that the candidates of a triplet prediction problem should be included in the training set of the base-level knowledge graph, which aligns with a realistic setting. By taking into account both the base-level knowledge graph and the higher-level triplets simultaneously, the problem becomes a transductive setting for \ours. This shows that simply converting the higher-level triplets into $G_T$ cannot replace our model. 

Table~\ref{tb:tp} shows the results of triplet prediction. We see that \oursq and \oursb significantly outperform all other state-of-the-art baseline methods in terms of all the three metrics on all three real-world datasets. Note that the number of candidates of a triplet prediction problem is equal to the number of base-level triplets in $\sE_{\text{train}}$. Therefore, achieving the MR of 18.7 on \fbh, for example, is surprising because we have 248,095 candidates in $\sE_{\text{train}}$. We visualize the embedding vectors generated by \oursq on \fbh in Figure~\ref{fig:tri_emb}. We take all higher-level triplets in the form of $\langle T_i$, ImpliesProfession, $T_j\rangle$ and visualize the embedding vectors of $T_i$ and $T_j$ using Principal Component Analysis. In Figure~\ref{fig:tri_emb}, we only highlight the base-level triplets $T_i$ and $T_j$ where $T_i$ is (Person~A, ActsIn, Film B) and $T_j$ is (Person~A, IsA, Actor). We see that both $\vT_i$ and $\vT_j$ embedding vectors are well-clustered, meaning that \ours generates embeddings by appropriately reflecting the structure of the higher-level triplets.

\begin{table*}
\small
\centering
\setlength{\tabcolsep}{0.42em}
\begin{tabular}{cccccccccc}
\toprule
 & \multicolumn{3}{c}{\fbh} & \multicolumn{3}{c}{\fbhe} & \multicolumn{3}{c}{\dbhe} \\
 & MR ($\downarrow$) & MRR ($\uparrow$) & Hit@10 ($\uparrow$) & MR ($\downarrow$) & MRR ($\uparrow$) & Hit@10 ($\uparrow$) & MR ($\downarrow$) & MRR ($\uparrow$) & Hit@10 ($\uparrow$) \\
\midrule
\aser & 1183.9{\scriptsize$\pm$0.0} & 0.251{\scriptsize$\pm$0.000} & 0.316{\scriptsize$\pm$0.000} & 970.7{\scriptsize$\pm$0.0} & 0.289{\scriptsize$\pm$0.000} & 0.382{\scriptsize$\pm$0.000} & 1893.5{\scriptsize$\pm$0.0} & 0.225{\scriptsize$\pm$0.000} & 0.348{\scriptsize$\pm$0.000} \\
\mnv & 3817.8{\scriptsize$\pm$58.9} & 0.328{\scriptsize$\pm$0.013} & 0.415{\scriptsize$\pm$0.009} & 3018.5{\scriptsize$\pm$45.8} & 0.407{\scriptsize$\pm$0.013} & 0.492{\scriptsize$\pm$0.014} & 2934.1{\scriptsize$\pm$32.2} & 0.362{\scriptsize$\pm$0.007} & 0.433{\scriptsize$\pm$0.014} \\
\mhop & 1878.2{\scriptsize$\pm$12.0} & 0.421{\scriptsize$\pm$0.003} & 0.578{\scriptsize$\pm$0.003} & 1447.3{\scriptsize$\pm$11.9} & 0.443{\scriptsize$\pm$0.002} & 0.615{\scriptsize$\pm$0.002} & 1012.3{\scriptsize$\pm$28.5} & 0.442{\scriptsize$\pm$0.007} & 0.652{\scriptsize$\pm$0.008} \\
\nelp & 185.9{\scriptsize$\pm$1.3} & 0.433{\scriptsize$\pm$0.002} & 0.648{\scriptsize$\pm$0.004} & 146.2{\scriptsize$\pm$1.0} & 0.466{\scriptsize$\pm$0.002} & 0.716{\scriptsize$\pm$0.007} & 32.2{\scriptsize$\pm$1.9} & 0.517{\scriptsize$\pm$0.006} & 0.756{\scriptsize$\pm$0.004} \\
\drum & 262.7{\scriptsize$\pm$13.3} & 0.394{\scriptsize$\pm$0.002} & 0.555{\scriptsize$\pm$0.003} & 207.6{\scriptsize$\pm$10.0} & 0.413{\scriptsize$\pm$0.010} & 0.620{\scriptsize$\pm$0.018} & 49.0{\scriptsize$\pm$3.9} & 0.470{\scriptsize$\pm$0.010} & 0.732{\scriptsize$\pm$0.012} \\
\anyb & 228.5{\scriptsize$\pm$11.8} & 0.380{\scriptsize$\pm$0.004} & 0.563{\scriptsize$\pm$0.013} & 166.0{\scriptsize$\pm$7.9} & 0.418{\scriptsize$\pm$0.002} & 0.607{\scriptsize$\pm$0.008} & 81.7{\scriptsize$\pm$4.0} & 0.403{\scriptsize$\pm$0.002} & 0.594{\scriptsize$\pm$0.004} \\
\ptranse & 214.8{\scriptsize$\pm$0.7} & 0.440{\scriptsize$\pm$0.001} & 0.686{\scriptsize$\pm$0.002} & 167.0{\scriptsize$\pm$1.8} & 0.516{\scriptsize$\pm$0.001} & 0.752{\scriptsize$\pm$0.001} & 19.3{\scriptsize$\pm$0.2} & 0.505{\scriptsize$\pm$0.004} & 0.780{\scriptsize$\pm$0.001} \\
\rpje & 212.5{\scriptsize$\pm$0.1} & 0.440{\scriptsize$\pm$0.001} & 0.686{\scriptsize$\pm$0.001} & 159.0{\scriptsize$\pm$0.0} & 0.528{\scriptsize$\pm$0.001} & 0.753{\scriptsize$\pm$0.001} & 19.3{\scriptsize$\pm$0.1} & 0.504{\scriptsize$\pm$0.004} & 0.779{\scriptsize$\pm$0.002} \\
\transd & 190.1{\scriptsize$\pm$18.0} & 0.300{\scriptsize$\pm$0.003} & 0.496{\scriptsize$\pm$0.005} & 165.6{\scriptsize$\pm$8.0} & 0.363{\scriptsize$\pm$0.003} & 0.529{\scriptsize$\pm$0.006} & 35.5{\scriptsize$\pm$1.0} & 0.436{\scriptsize$\pm$0.006} & 0.708{\scriptsize$\pm$0.005} \\
\anal & 341.0{\scriptsize$\pm$218.7} & 0.182{\scriptsize$\pm$0.065} & 0.291{\scriptsize$\pm$0.125} & 113.3{\scriptsize$\pm$2.0} & 0.409{\scriptsize$\pm$0.004} & 0.581{\scriptsize$\pm$0.004} & 279.1{\scriptsize$\pm$197.1} & 0.140{\scriptsize$\pm$0.089} & 0.253{\scriptsize$\pm$0.166} \\
\quate & 163.7{\scriptsize$\pm$3.6} & 0.346{\scriptsize$\pm$0.006} & 0.494{\scriptsize$\pm$0.011} & 1546.4{\scriptsize$\pm$98.0} & 0.124{\scriptsize$\pm$0.022} & 0.189{\scriptsize$\pm$0.014} & 551.6{\scriptsize$\pm$40.5} & 0.208{\scriptsize$\pm$0.013} & 0.309{\scriptsize$\pm$0.023} \\
\bique & 111.0{\scriptsize$\pm$0.9} & 0.423{\scriptsize$\pm$0.002} & 0.641{\scriptsize$\pm$0.002} & 90.1{\scriptsize$\pm$0.5} & 0.387{\scriptsize$\pm$0.009} & 0.617{\scriptsize$\pm$0.011} & 29.5{\scriptsize$\pm$1.2} & 0.378{\scriptsize$\pm$0.007} & 0.677{\scriptsize$\pm$0.004} \\
\oursq & \underline{7.0{\scriptsize$\pm$0.3}} & \underline{0.752{\scriptsize$\pm$0.005}} & \underline{0.906{\scriptsize$\pm$0.002}} & \textbf{11.0{\scriptsize$\pm$0.3}} & \textbf{0.698{\scriptsize$\pm$0.004}} & \textbf{0.839{\scriptsize$\pm$0.003}} & \underline{12.5{\scriptsize$\pm$1.0}} & \underline{0.606{\scriptsize$\pm$0.009}} & \underline{0.828{\scriptsize$\pm$0.010}} \\
\oursb & \textbf{6.6{\scriptsize$\pm$0.3}} & \textbf{0.762{\scriptsize$\pm$0.007}} & \textbf{0.911{\scriptsize$\pm$0.002}} & \underline{12.8{\scriptsize$\pm$0.4}} & \underline{0.696{\scriptsize$\pm$0.005}} & \underline{0.834{\scriptsize$\pm$0.002}} & \textbf{3.2{\scriptsize$\pm$0.1}} & \textbf{0.801{\scriptsize$\pm$0.003}} & \textbf{0.958{\scriptsize$\pm$0.002}} \\
\bottomrule
\end{tabular}
\vspace{-0.2cm}
\caption{Results of Conditional Link Prediction. The best scores are boldfaced and the second best scores are underlined. Our models, \oursq and \oursb, significantly outperform all baseline methods in terms of all metrics on all datasets.}
\label{tb:clp}
\end{table*}

\begin{table*}[t]
\scriptsize
\centering
\begin{tabular}{lcl}
\toprule
Problem & \hspace{4.6em} & Prediction by \oursq \\
\midrule
$\langle$(\texttt{?}, HasAFriendshipWith, Kelly\_Preston), EquivalentTo, (Kelly\_Preston, HasAFriendshipWith, George\_Clooney)$\rangle$ & & George\_Clooney \\
\cdashline{1-3}
$\langle$(\texttt{?}, HasAFriendshipWith, Kelly\_Preston), EquivalentTo, (Kelly\_Preston, HasAFriendshipWith, Tom\_Cruise)$\rangle$ & & Tom\_Cruise \\
\midrule
$\langle$(Joe\_Jonas, IsA, \texttt{?}), ImpliesProfession, (Joe\_Jonas, IsA, Actor)$\rangle$ & & Voice\_Actor \\
\cdashline{1-3}
$\langle$(Joe\_Jonas, IsA, \texttt{?}), ImpliesProfession, (Joe\_Jonas, IsA, Musician)$\rangle$ & & Singer-songwriter \\
\midrule
$\langle$(Bucknell\_University, HasAHeadquarterIn, Pennsylvania), ImpliesLocation, (\texttt{?}, Contains, Bucknell\_University)$\rangle$ & & Pennsylvania \\
\cdashline{1-3}
$\langle$(Bucknell\_University, HasAHeadquarterIn, United\_States), ImpliesLocation, (\texttt{?}, Contains, Bucknell\_University)$\rangle$ & & United\_States \\
\midrule 
$\langle$(Saturn\_Award\_for\_Best\_Director, Nominates, Avatar), PrerequisiteFor, (Avatar, Wins, \texttt{?})$\rangle$ & & Saturn\_Award\_for\_Best\_Director \\
\cdashline{1-3}
$\langle$(Academy\_Award\_for\_Best\_Visual\_Effects, Nominates, Avatar), PrerequisiteFor, (Avatar, Wins, \texttt{?})$\rangle$ & & Academy\_Award\_for\_Best\_Visual\_Effects \\
\bottomrule
\end{tabular}
\vspace{-0.2cm}
\caption{Examples of Conditional Link Prediction on \fbhe. \ours correctly predicts the answers for all the above problems.}
\label{tb:qa}
\vspace{-0.3cm}
\end{table*}

\begin{table}[t]
\scriptsize
\centering
\setlength{\tabcolsep}{0.76em}
\begin{tabular}{ccccc}
\toprule
 & \multicolumn{2}{c}{\fbhe} & \multicolumn{2}{c}{\dbhe} \\
 & MR ($\downarrow$) & Hit@10 ($\uparrow$) & MR ($\downarrow$) & Hit@10 ($\uparrow$) \\
\midrule
\aser & 1489.3{\scriptsize$\pm$0.0} & 0.323{\scriptsize$\pm$0.000} & 2218.8{\scriptsize$\pm$0.0} & 0.197{\scriptsize$\pm$0.000}\\
\mnv & 3828.4{\scriptsize$\pm$56.9} & 0.339{\scriptsize$\pm$0.003} & 3530.7{\scriptsize$\pm$50.1} & 0.297{\scriptsize$\pm$0.006}\\
\mhop & 2284.0{\scriptsize$\pm$9.5} & 0.500{\scriptsize$\pm$0.001} & 2489.4{\scriptsize$\pm$15.3} & 0.404{\scriptsize$\pm$0.004}\\
\nelp & 1942.5{\scriptsize$\pm$0.5} & 0.486{\scriptsize$\pm$0.001} & 2904.8{\scriptsize$\pm$0.6} & 0.357{\scriptsize$\pm$0.001}\\
\drum & 1945.6{\scriptsize$\pm$0.8} & 0.490{\scriptsize$\pm$0.002} & 2904.7{\scriptsize$\pm$0.7} & 0.359{\scriptsize$\pm$0.001}\\
\anyb & 342.0{\scriptsize$\pm$4.6} & 0.526{\scriptsize$\pm$0.002} & 879.1{\scriptsize$\pm$5.7} & 0.364{\scriptsize$\pm$0.003}\\
\ptranse & 2077.6{\scriptsize$\pm$10.3} & 0.333{\scriptsize$\pm$0.000} & 3346.0{\scriptsize$\pm$20.0} & 0.277{\scriptsize$\pm$0.002}\\
\rpje & 1754.6{\scriptsize$\pm$7.5} & 0.368{\scriptsize$\pm$0.001} & 2991.7{\scriptsize$\pm$28.1} & 0.341{\scriptsize$\pm$0.000}\\
\transd & 166.3{\scriptsize$\pm$1.3} & 0.527{\scriptsize$\pm$0.001} & 429.0{\scriptsize$\pm$7.5} & 0.423{\scriptsize$\pm$0.001} \\
\anal & 227.3{\scriptsize$\pm$8.3} & 0.486{\scriptsize$\pm$0.002} & 621.5{\scriptsize$\pm$20.9} & 0.323{\scriptsize$\pm$0.008}\\
\quate & 139.0{\scriptsize$\pm$1.6} & 0.581{\scriptsize$\pm$0.001} & 409.6{\scriptsize$\pm$8.5} & 0.440{\scriptsize$\pm$0.001}\\
\bique & 134.9{\scriptsize$\pm$0.9} & 0.583{\scriptsize$\pm$0.001} & \textbf{376.6{\scriptsize$\pm$3.5}} & \textbf{0.446{\scriptsize$\pm$0.002}}\\
\oursq & \underline{125.2{\scriptsize$\pm$0.9}} & \underline{0.584{\scriptsize$\pm$0.001}} & 405.4{\scriptsize$\pm$4.1} & 0.438{\scriptsize$\pm$0.002}\\
\oursb & \textbf{123.5{\scriptsize$\pm$1.0}} & \textbf{0.586{\scriptsize$\pm$0.001}} & \underline{377.3{\scriptsize$\pm$6.7}} & \underline{0.444{\scriptsize$\pm$0.001}}\\
\bottomrule
\end{tabular}
\vspace{-0.2cm}
\caption{Results of Base-Level Link Prediction.}
\label{tb:blp}
\vspace{-0.7cm}
\end{table}
\vspace{-0.2cm}
\subsection{Conditional Link Prediction}
\vspace{-0.2cm}
To solve a conditional link prediction problem, \ours uses the scoring function $F_{\text{clp}}(x)$. On the other hand, the baseline methods cannot directly solve the conditional link prediction problem. To allow the baseline methods to solve $\langle T_i, \rhat, (h_j,r_j,?) \rangle$\footnote{We consider all four problems by changing the position of ?.}, we define a scoring function of the baseline methods as follows: $F(x)\coloneqq f(\vh_j,\vr_j,\vx) + f(\vT_i,\vrhat,z(h_j,r_j,x))$ for all $x\in\sV$ where the former is computed on the original base-level knowledge graph, the latter is computed on $G_T$, $z(h_j,r_j,x)$ returns an embedding vector of $(h_j,r_j,x)$, and $f(\cdot)$ is the scoring function of each baseline method. We cannot get $f(\vT_i,\vrhat,z(h_j,r_j,x))$ if $(h_j,r_j,x)\notin\sE_{\text{train}}$. In that case, we compute the score using the randomly initialized vectors in \ptranse, \rpje, \transd, \anal, \quate, and \bique, whereas we set $f(\vT_i,\vrhat,z(h_j,r_j,x))=0$ for the other baseline methods by considering the mechanisms of how each of the baseline methods assigns scores. In Table~\ref{tb:clp}, \oursq and \oursb significantly outperform all other baseline methods in conditional link prediction on all three real-world datasets. In Table~\ref{tb:qa}, we show some example problems of conditional link prediction in \fbhe and the predictions made by \oursq where it correctly predicts the answers. When we consider a problem in the form of $\langle T_i, \rhat, (h_j,r_j,?) \rangle$, even though we have the same problem of $(h_j,r_j,?)$, the answer becomes different depending on $T_i$. This is the difference between the typical base-level link prediction and the conditional link prediction.

\begin{table*}[t]
\centering
\scriptsize
\setlength{\tabcolsep}{0.4em}
\begin{tabular}{cllccl}
\toprule
 & Relation Sequence $p_k$ & Relation $r$ & $c(p_k, r)$ & $|S_{kr}|$ & Examples of the Augmented Triplets \\
\midrule
\multirow{5}{*}{\fbhe} & NominatesIn, $\mbox{NominatesIn}^{-1}$, ActsIn, \textbf{ImpliesProfession}, IsA & IsA & 0.86 & 610 & (Patty\_Duke, IsA, Actor) \\
 & ParticipatesIn, $\mbox{ParticipatesIn}^{-1}$, \textbf{ImpliesSports}, $\mbox{Plays}^{-1}$, ParticipatesIn & ParticipatesIn & 0.81 & 57 & (Houston\_Rockets, ParticipatesIn, 2003\_NBA\_Draft) \\
 & Plays, $\mbox{Plays}^{-1}$, $\mbox{\textbf{ImpliesSports}}^{-1}$, HasPosition & HasPosition & 0.78 & 295 & (Bayer\_04\_Leverkusen, HasPosition, Forward) \\
 & Contains, $\mbox{Contains}^{-1}$, $\mbox{\textbf{ImpliesLocation}}^{-1}$, HasAHeadquarterIn & Contains & 0.72 & 81 & (United\_States, Contains, Charlottesville\_Virginia) \\
 & $\mbox{Program}^{-1}$, Program, Language & Language & 0.70 & 148 & (David\_Copperfield\_(Film), Language, English\_Language) \\
\cdashline{1-6}
\multirow{5}{*}{\dbhe} & Genre, $\mbox{\textbf{ImpliesGenre}}^{-1}$, Genre, $\mbox{Genre}^{-1}$, $\mbox{\textbf{ImpliesGenre}}^{-1}$, Genre & Genre & 0.78 & 120 & (Kenny\_Rogers, Genre, Pop\_Rock) \\
 & IsPartOf, IsPartOf, \textbf{ImpliesLocation}, IsPartOf & IsPartOf & 0.76 & 69 & (San\_Pedro\_Los\_Angeles, IsPartOf, California) \\
 & IsPartOf, $\mbox{IsPartOf}^{-1}$, $\mbox{\textbf{ImpliesLocation}}^{-1}$, $\mbox{IsPartOf}^{-1}$, TimeZone & TimeZone & 0.75 & 122 & (Brockton\_Massachusetts, TimeZone, Eastern\_Time\_Zone) \\
 & $\mbox{IsProducedBy}^{-1}$, IsProducedBy, \textbf{ImpliesProfession}, IsA & IsA &  0.73 & 80 & (Jim\_Wilson, IsA, Film\_Producer) \\
 & Region, $\mbox{Region}^{-1}$, Country & Country & 0.70 & 41 & (Pontefract, Country, England)\\
\bottomrule
\end{tabular}
\caption{Examples of the Augmented Triplets in \fbhe and \dbhe. The higher-level relations are boldfaced.}
\label{tb:aug2}
\vspace{-0.2cm}
\end{table*}

\begin{table}[t]
\small
\centering
\setlength{\tabcolsep}{0.44em}
\begin{tabular}{lccc}
\toprule
 & \fbh & \fbhe & \dbhe \\
\midrule
No. of unique $(p_k,r)$ & 340,194 & 349,120 & 149,365 \\
No. of $(p_k,r)$ with $c(p_k,r)\geq 0.7$ & 35,803 & 39,727 & 7,030 \\
No. of augmented triplets & 16,601 & 17,463 & 2,026 \\
$|\sS\cap\sE_{\text{valid}}|+|\sS\cap\sE_{\text{test}}|$ & 5,237 & 5,380 & 316 \\
\bottomrule
\end{tabular}
\vspace{-0.2cm}
\caption{Statistics of the Augmented Triplets.}
\label{tb:aug}
\end{table}

\begin{table}[t]
\small
\centering
\begin{tabular}{ccccc}
\toprule
 & & \fbh & \fbhe & \dbhe \\
\midrule
\multirow{2}{*}{TP} & $L_{\text{base}} + L_{\text{high}}$ & 19.2 & 28.1 & 65.4 \\
& $L_{\text{base}} + L_{\text{high}}+ L_{\text{aug}}$ & 18.7 & 33.1 & 56.6 \\
\cdashline{1-5}
\multirow{2}{*}{CLP} & $L_{\text{base}} + L_{\text{high}}$ & 8.3 & 12.5 & 12.4 \\
& $L_{\text{base}} + L_{\text{high}}+ L_{\text{aug}}$ & 7.0 & 11.0 & 12.5 \\
\cdashline{1-5}
\multirow{4}{*}{BLP} & $L_{\text{base}}$ & 139.0 & 139.0 & 409.6 \\
& $L_{\text{base}} + L_{\text{high}}$ & 138.4 & 138.4 & 408.1 \\
& $L_{\text{base}} + L_{\text{aug}}$ & 124.7 & 125.2 & 404.9 \\
& $L_{\text{base}} + L_{\text{high}}+ L_{\text{aug}}$ & 124.7 & 125.2 & 405.4 \\
\bottomrule
\end{tabular}
\caption{Ablation study of \ours with different combinations of the loss terms. The average MR scores on triplet prediction (TP), conditional link prediction (CLP), and the base-level link prediction (BLP).}
\label{tb:ab}
\end{table}

\begin{table}[t]
\scriptsize
\centering
\setlength{\tabcolsep}{0.62em}
\begin{tabular}{cccccccc}
\toprule
 & & \multicolumn{3}{c}{Triplet Prediction} & \multicolumn{3}{c}{Conditional LP} \\
 $\rhat$ & Freq. & MR & MRR & Hit@10 & MR & MRR & Hit@10 \\
\midrule
EquivalentTo & 98 & 17.5 & 0.416 & 0.679 & 2.2 & 0.744 & 0.977 \\
ImpliesLanguage & 29 & 35.6 & 0.292 & 0.578 & 18.4 & 0.632 & 0.786 \\
ImpliesProfession & 210 & 71.3 & 0.427 & 0.569 & 11.5 & 0.704 & 0.844 \\
ImpliesLocation & 163 & 42.2 & 0.219 & 0.463 & 9.4 & 0.502 & 0.816 \\
ImpliesTimeZone & 44 & 20.6 & 0.354 & 0.631 & 17.8 & 0.604 & 0.707 \\
ImpliesGenre & 84 & 113.8 & 0.177 & 0.345 & 32.6 & 0.408 & 0.681 \\
NextAlmaMater & 14 & 71.0 & 0.161 & 0.379 & 2.5 & 0.651 & 0.971 \\
TransfersTo & 29 & 67.0 & 0.140 & 0.374 & 5.7 & 0.527 & 0.537 \\
\bottomrule
\end{tabular}
\caption{Performance of \ours per higher-level relation in \dbhe. Freq. indicates the number of higher-level triplets in $\sH_{\text{test}}$ associated with $\rhat$.}
\label{tb:rel}
\end{table}

\vspace{-0.2cm}
\subsection{Base-Level Link Prediction}
\vspace{-0.2cm}
We present the performance of the typical base-level link prediction in Table~\ref{tb:blp}. Since the base-level knowledge graphs of \fbh and \fbhe are identical, the performance of all baseline methods is the same on \fbh and \fbhe. The base-level link prediction performance of \ours on \fbh and \fbhe is also very similar to each other. We observed that the MRR scores of our \ours models and the two best baselines are almost the same on \fbhe and \dbhe. On \fbhe, the average MRR scores of \oursq and \quate are both 0.354, and those of \oursb and \bique are both 0.356. On \dbhe, the average MRR score of \oursq is 0.265 and that of \quate is 0.264; the average MRR score of \oursb is 0.275 and that of \bique is 0.274. Overall, our \ours models show comparable results to the baseline methods for the typical link prediction task; our \ours models have the extra capability of dealing with the triplet prediction and conditional link prediction tasks.

\vspace{-0.2cm}
\subsection{Data Augmentation of \ours}
\vspace{-0.2cm}
We analyze the augmented triplets that are added by our data augmentation strategy. In Table~\ref{tb:aug2}, we show some examples of a relation sequence $p_k$, a relation $r$, and the confidence of $(p_k,r)$, the number of augmented triplets based on $(p_k,r)$ denoted by $|\sS_{kr}|$, and examples of the augmented triplets in \fbhe and \dbhe. According to our random walk policy, we do not allow going back to an entity that has already been visited. Thus, even though a relation and its reverse relation are consecutively appeared in a relation sequence in Table~\ref{tb:aug2}, it does not mean that we return back to the previous entity; instead, it means that the walk steps another entity adjacent to the corresponding relation. In Table~\ref{tb:aug}, we show statistics of the augmented triplets. Among the diverse combinations of a relation sequence $p_k$ and a relation $r$, we consider the $(p_k,r)$ pairs whose confidence scores are greater than or equal to $0.7$. It is interesting to see that there exist considerable overlaps between the set $\sS$ of the augmented triplets and $\sE_{\text{valid}}$ and $\sE_{\text{test}}$, indicating that our augmented triplets include many ground-truth triplets that are missing in the training set.
\vspace{-0.2cm}
\subsection{Ablation Study of \ours}
\vspace{-0.2cm}
In \ours, we have three different types of loss terms: $L_{\text{base}}$, $L_{\text{high}}$, and $L_{\text{aug}}$. Using different combinations of these loss terms, we measure the performance of \ours to check the importance of each loss term. Table~\ref{tb:ab} shows the average MR scores of \oursq with different combinations of the loss terms in three tasks: triplet prediction (TP), conditional link prediction (CLP), and base-level link prediction (BLP). Note that TP and CLP require at least two terms, $L_{\text{base}}$ and $L_{\text{high}}$. Also, Table~\ref{tb:rel} shows the performance of \oursq per higher-level relation in \dbhe, where Freq. indicates the number of higher-level triplets in $\sH_{\text{test}}$ associated with $\rhat$. Among the eight higher-level relations in \dbhe, NextAlmaMater and TransfersTo require externally-sourced knowledge. While EquivalentTo is the easiest one, the performance on the other higher-level relations varies depending on the tasks and the metrics.

\vspace{-0.2cm}
\section{Conclusion and Future Work}
\vspace{-0.2cm}
We define a bi-level knowledge graph by introducing the higher-level relationships between triplets. We propose \ours, which takes into account the structures of the base-level triplets, the higher-level triplets, and the augmented triplets. Experimental results show that \ours significantly outperforms state-of-the-art methods in terms of the two newly defined tasks: triplet prediction and conditional link prediction. We believe our method can contribute to advancing many knowledge-based applications, including conditional QA~\cite{condqa} and multi-hop QA~\cite{hgn}, with a special emphasis on mixing a neural language model and structured knowledge~\cite{qagnn}.

We plan to analyze the pros and cons of our bi-level knowledge graphs and compare them with other forms of extended knowledge graphs, such as hyper-relational knowledge graphs~\cite{hynt}. Also, we will extend \ours and the proposed tasks to an inductive learning scenario where both entities and relations can be new at inference time~\cite{ingram}.

\vspace{-0.2cm}
\section{Acknowledgments}
\vspace{-0.2cm}
This research was supported by IITP grants funded by the Korean government MSIT 2022-0-00369, 2020-0-00153 (Penetration Security Testing of ML Model Vulnerabilities and Defense) and NRF of Korea funded by the Korean Government MSIT 2018R1A5A1059921, 2022R1A2C4001594.

\bibliographystyle{abbrv}
\bibliography{bive_ref}

\begin{thebibliography}{10}

\bibitem{dbp}
S.~Auer, C.~Bizer, G.~Kobilarov, J.~Lehmann, R.~Cyganiak, and Z.~Ives.
\newblock Dbpedia: A nucleus for a web of open data.
\newblock In {\em Proceedings of the 6th International Semantic Web Conference
  and the 2nd Asian Semantic Web Conference}, pages 722--735, 2007.

\bibitem{fbl}
K.~Bollacker, C.~Evans, P.~Paritosh, T.~Sturge, and J.~Taylor.
\newblock Freebase: A collaboratively created graph database for structuring
  human knowledge.
\newblock In {\em Proceedings of the 2008 ACM SIGMOD International Conference
  on Management of Data}, pages 1247--1250, 2008.

\bibitem{transe}
A.~Bordes, N.~Usunier, A.~Garcia-Dur\'{a}n, J.~Weston, and O.~Yakhnenko.
\newblock Translating embeddings for modeling multi-relational data.
\newblock In {\em Proceedings of the International Conference on Neural
  Information Processing Systems}, page 2787–2795, 2013.

\bibitem{atth}
I.~Chami, A.~Wolf, D.-C. Juan, F.~Sala, S.~Ravi, and C.~Ré.
\newblock Low-dimensional hyperbolic knowledge graph embeddings.
\newblock In {\em Proceedings of the 58th Annual Meeting of the Association for
  Computational Linguistics}, pages 6901--6914, 2020.

\bibitem{hynt}
C.~Chung, J.~Lee, and J.~J. Whang.
\newblock Representation learning on hyper-relational and numeric knowledge
  graphs with transformers.
\newblock In {\em Proceedings of the 29th ACM SIGKDD Conference on Knowledge
  Discovery and Data Mining}, pages 310--322, 2023.

\bibitem{meta}
C.~Chung and J.~J. Whang.
\newblock Knowledge graph embedding via metagraph learning.
\newblock In {\em Proceedings of the 44th International ACM SIGIR Conference on
  Research and Development in Information Retrieval}, pages 2212--2216, 2021.

\bibitem{mnv}
R.~Das, S.~Dhuliawala, M.~Zaheer, L.~Vilnis, I.~Durugkar, A.~Krishnamurthy,
  A.~Smola, and A.~McCallum.
\newblock Go for a walk and arrive at the answer: Reasoning over paths in
  knowledge bases using reinforcement learning.
\newblock In {\em Proceedings of the 6th International Conference on Learning
  Representations}, 2018.

\bibitem{inject}
T.~Demeester, T.~Rocktäschel, and S.~Riedel.
\newblock Lifted rule injection for relation embeddings.
\newblock In {\em Proceedings of the 2016 Conference on Empirical Methods in
  Natural Language Processing}, pages 1389--1399, 2016.

\bibitem{hgn}
Y.~Fang, S.~Sun, Z.~Gan, R.~Pillai, S.~Wang, and J.~Liu.
\newblock Hierarchical graph network for multi-hop question answering.
\newblock In {\em Proceedings of the 2020 Conference on Empirical Methods in
  Natural Language Processing}, pages 8823--8838, 2020.

\bibitem{ttv}
V.~Fionda and G.~Pirrò.
\newblock Learning triple embeddings from knowledge graphs.
\newblock In {\em Proceedings of the 34th AAAI Conference on Artificial
  Intelligence}, pages 3874--3881, 2020.

\bibitem{kblrn}
A.~Garcia-Duran and M.~Niepert.
\newblock Kblrn: End-to-end learning of knowledge base representations with
  latent, relational, and numerical features.
\newblock In {\em Proceedings of the 34th Conference on Uncertainty in
  Artificial Intelligence}, pages 372--381, 2018.

\bibitem{bique}
J.~Guo and S.~Kok.
\newblock Bique: Biquaternionic embeddings of knowledge graphs.
\newblock In {\em Proceedings of the 2021 Conference on Empirical Methods in
  Natural Language Processing}, page 8338–8351, 2021.

\bibitem{kale}
S.~Guo, Q.~Wang, L.~Wang, B.~Wang, and L.~Guo.
\newblock Jointly embedding knowledge graphs and logical rules.
\newblock In {\em Proceedings of the 2016 Conference on Empirical Methods in
  Natural Language Processing}, pages 192--202, 2016.

\bibitem{openke}
X.~Han, S.~Cao, X.~Lv, Y.~Lin, Z.~Liu, M.~Sun, and J.~Li.
\newblock {O}pen{KE}: An open toolkit for knowledge embedding.
\newblock In {\em Proceedings of the 2018 Conference on Empirical Methods in
  Natural Language Processing: System Demonstrations}, pages 139--144, 2018.

\bibitem{joie}
J.~Hao, M.~Chen, W.~Yu, Y.~Sun, and W.~Wang.
\newblock Universal representation learning of knowledge bases by jointly
  embedding instances and ontological concepts.
\newblock In {\em Proceedings of the 25th ACM SIGKDD International Conference
  on Knowledge Discovery and Data Mining}, page 1709–1719, 2019.

\bibitem{transd}
G.~Ji, S.~He, L.~Xu, K.~Liu, and J.~Zhao.
\newblock Knowledge graph embedding via dynamic mapping matrix.
\newblock In {\em Proceedings of the 53rd Annual Meeting of the Association for
  Computational Linguistics and the 7th International Joint Conference on
  Natural Language Processing}, pages 687--696, 2015.

\bibitem{kgss}
S.~Ji, S.~Pan, E.~Cambria, P.~Marttinen, and P.~S. Yu.
\newblock A survey on knowledge graphs: Representation, acquisition and
  applications.
\newblock {\em IEEE Transactions on Neural Networks and Learning Systems},
  33(2):494--514, 2022.

\bibitem{ptd}
Y.~Jiang, X.~Wang, H.~Fan, Q.~Liu, B.~Du, and H.~Zhu.
\newblock Modeling relation path for knowledge graph via dynamic projection.
\newblock In {\em The 32nd International Conference on Software Engineering and
  Knowledge Engineering}, pages 65--70, 2020.

\bibitem{literale}
A.~Kristiadi, M.~A. Khan, D.~Lukovnikov, J.~Lehmann, and A.~Fischer.
\newblock Incorporating literals into knowledge graph embeddings.
\newblock In {\em Proceedings of the 18th International Semantic Web
  Conference}, pages 347--363, 2019.

\bibitem{robo}
J.~H. Kwak, J.~Lee, J.~J. Whang, and S.~Jo.
\newblock Semantic grasping via a knowledge graph of robotic manipulation: A
  graph representation learning approach.
\newblock {\em IEEE Robotics and Automation Letters}, 7(4):9397--9404, 2022.

\bibitem{pra}
N.~Lao and W.~W. Cohen.
\newblock Relational retrieval using a combination of path-constrained random
  walks.
\newblock {\em Machine Learning}, 81:53--67, 2010.

\bibitem{ingram}
J.~Lee, C.~Chung, and J.~J. Whang.
\newblock {I}n{G}ram: Inductive knowledge graph embedding via relation graphs.
\newblock In {\em Proceedings of the 40th International Conference on Machine
  Learning}, pages 18796--18809, 2023.

\bibitem{daug}
G.~Li, Z.~Sun, L.~Qian, Q.~Guo, and W.~Hu.
\newblock Rule-based data augmentation for knowledge graph embedding.
\newblock {\em AI Open}, 2:186--196, 2021.

\bibitem{mhop}
X.~V. Lin, R.~Socher, and C.~Xiong.
\newblock Multi-hop knowledge graph reasoning with reward shaping.
\newblock In {\em Proceedings of the 2018 Conference on Empirical Methods in
  Natural Language Processing}, pages 3243--3253, 2018.

\bibitem{pte}
Y.~Lin, Z.~Liu, H.~Luan, M.~Sun, S.~Rao, and S.~Liu.
\newblock Modeling relation paths for representation learning of knowledge
  bases.
\newblock In {\em Proceedings of the 2015 Conference on Empirical Methods in
  Natural Language Processing}, pages 705--714, 2015.

\bibitem{anal}
H.~Liu, Y.~Wu, and Y.~Yang.
\newblock Analogical inference for multi-relational embeddings.
\newblock In {\em Proceedings of the 34th International Conference on Machine
  Learning}, page 2168–2178, 2017.

\bibitem{anyb}
C.~Meilicke, M.~W. Chekol, D.~Ruffinelli, and H.~Stuckenschmidt.
\newblock Anytime bottom-up rule learning for knowledge graph completion.
\newblock In {\em Proceedings of the 28th International Joint Conference on
  Artificial Intelligence}, pages 3137--3143, 2019.

\bibitem{lenn}
M.~Nayyeri, C.~Xu, M.~M. Alam, J.~Lehmann, and H.~S. Yazdi.
\newblock Logicenn: A neural based knowledge graphs embedding model with
  logical rules.
\newblock {\em IEEE Transactions on Pattern Analysis and Machine Intelligence},
  2021.

\bibitem{rpje}
G.~Niu, Y.~Zhang, B.~Li, P.~Cui, S.~Liu, J.~Li, and X.~Zhang.
\newblock Rule-guided compositional representation learning on knowledge
  graphs.
\newblock In {\em Proceedings of the 34th AAAI Conference on Artificial
  Intelligence}, pages 2950--2958, 2020.

\bibitem{hinge}
P.~Rosso, D.~Yang, and P.~Cudr\'{e}-Mauroux.
\newblock Beyond triplets: Hyper-relational knowledge graph embedding for link
  prediction.
\newblock In {\em Proceedings of The Web Conference 2020}, page 1885–1896,
  2020.

\bibitem{drum}
A.~Sadeghian, M.~Armandpour, P.~Ding, and D.~Z. Wang.
\newblock Drum: End-to-end differentiable rule mining on knowledge graphs.
\newblock In {\em Proceedings of the 32nd International Conference on Neural
  Information Processing Systems}, pages 15347--15357, 2018.

\bibitem{atomic}
M.~Sap, R.~L. Bras, E.~Allaway, C.~Bhagavatula, N.~Lourie, H.~Rashkin, B.~Roof,
  N.~A. Smith, and Y.~Choi.
\newblock Atomic: An atlas of machine commonsense for if-then reasoning.
\newblock In {\em Proceedings of the 33rd AAAI Conference on Artificial
  Intelligence}, pages 3027--3035, 2019.

\bibitem{condqa}
H.~Sun, W.~Cohen, and R.~Salakhutdinov.
\newblock Conditionalqa: A complex reading comprehension dataset with
  conditional answers.
\newblock In {\em Proceedings of the 60th Annual Meeting of the Association for
  Computational Linguistics}, pages 3627--3637, 2022.

\bibitem{fb}
K.~Toutanova and D.~Chen.
\newblock Observed versus latent features for knowledge base and text
  inference.
\newblock In {\em Proceedings of the 3rd Workshop on Continuous Vector Space
  Models and their Compositionality}, pages 57--66, 2015.

\bibitem{cvt}
N.~Voskarides, E.~Meij, R.~Reinanda, A.~Khaitan, M.~Osborne, G.~Stefanoni,
  P.~Kambadur, and M.~de~Rijke.
\newblock Weakly-supervised contextualization of knowledge graph facts.
\newblock In {\em Proceedings of the 41st International ACM SIGIR Conference on
  Research and Development in Information Retrieval}, pages 765--774, 2018.

\bibitem{kgs}
Q.~Wang, Z.~Mao, B.~Wang, and L.~Guo.
\newblock Knowledge graph embedding: A survey of approaches and applications.
\newblock {\em IEEE Transactions on Knowledge and Data Engineering},
  29(12):2724--2743, 2017.

\bibitem{transea}
Y.~Wu and Z.~Wang.
\newblock Knowledge graph embedding with numeric attributes of entities.
\newblock In {\em Proceedings of the 3rd Workshop on Representation Learning
  for {NLP}}, pages 132--136, 2018.

\bibitem{nell}
W.~Xiong, T.~Hoang, and W.~Y. Wang.
\newblock Deeppath: A reinforcement learning method for knowledge graph
  reasoning.
\newblock In {\em Proceedings of the Conference on Empirical Methods in Natural
  Language Processing}, pages 564--573, 2017.

\bibitem{nelp}
F.~Yang, Z.~Yang, and W.~W. Cohen.
\newblock Differentiable learning of logical rules for knowledge base
  reasoning.
\newblock In {\em Proceedings of the 31st International Conference on Neural
  Information Processing Systems}, pages 2316--2325, 2017.

\bibitem{qagnn}
M.~Yasunaga, H.~Ren, A.~Bosselut, P.~Liang, and J.~Leskovec.
\newblock Qa-gnn: Reasoning with language models and knowledge graphs for
  question answering.
\newblock In {\em Proceedings of the 2021 Conference of the North American
  Chapter of the Association for Computational Linguistics: Human Language
  Technologies}, pages 535--546, 2021.

\bibitem{aser}
H.~Zhang, X.~Liu, H.~Pan, Y.~Song, and C.~W.-K. Leung.
\newblock Aser: A large-scale eventuality knowledge graph.
\newblock In {\em Proceedings of The Web Conference 2020}, pages 201--211,
  2020.

\bibitem{quate}
S.~Zhang, Y.~Tay, L.~Yao, and Q.~Liu.
\newblock Quaternion knowledge graph embeddings.
\newblock In {\em Proceedings of the 33rd International Conference on Neural
  Information Processing Systems}, pages 2735--2745, 2019.

\end{thebibliography}
\clearpage

\onecolumn
\section{\Large{Appendix}}
\vspace{-0.2cm}
\section{Experimental Settings}
\vspace{-0.2cm}
All experiments are conducted on machines equipped with Intel(R) Xeon(R) E5-2690 v4 CPUs and 512GB memory. We use RTX A6000 GPUs unless otherwise stated.

\vspace{-0.2cm}
\subsection{Baseline Methods}
\vspace{-0.2cm}
In experiments, we use 12 different baseline methods which are \aser~\cite{aser}, \mnv~\cite{mnv}, \mhop~\cite{mhop}, \nelp~\cite{nelp}, \drum~\cite{drum}, \anyb~\cite{anyb}, \ptranse~\cite{pte}, \rpje~\cite{rpje}, \transd~\cite{transd}, \anal~\cite{anal}, \quate~\cite{quate}, and \bique~\cite{bique}. We use GeForce RTX 2080Ti GPUs to run \mnv, \nelp, and \drum because these methods use TensorFlow version 1.

For \aser, we implement the string matching based inference model described in~\cite{aser}. The original program of \mnv is designed only to predict a tail entity. We add reversed triplets to also predict a head entity. In \mhop, we set all the add\_reverse\_edge options to be true. We run \anyb with different rule learning times: 10 seconds, 100 seconds, 1,000 seconds and 10,000 seconds. Since the results of running 10,000 seconds are similar to those of 1,000 seconds, we use the results of 1,000 seconds. We use the best hyperparameters provided by the authors of each of the baseline methods except \ptranse, \rpje, \transd, \anal, \quate, and \bique; we tune the hyperparameters of \ptranse, \rpje, \transd, \anal, \quate, and \bique because the best hyperparameters are not provided for these methods.

For \ptranse and \rpje, we tune the learning rate $\alpha$ using the range of $\alpha = \{ 10^{-6}, 5 \cdot 10^{-6}, 10^{-5}, 5 \cdot 10^{-5}, 10^{-4}, 5 \cdot 10^{-4} \}$ and the margin $\gamma$ using the range of $\gamma = \{ 0.5, 1.0, 2.0 \}$. We use \transd, \anal, \quate, and \bique, which are implemented based on OpenKE~\cite{openke}. For \transd, we tune the learning rate $\alpha$ using the range of $\alpha = \{ 0.1, 0.5, 1.0, 2.0, 5.0 \}$ and the margin $\gamma$ using the range of $\gamma = \{ 2.0, 5.0, 10.0 \}$. For \anal, we tune the learning rate $\alpha$ using the range of $\alpha = \{ 0.001, 0.005, 0.01, 0.05, 0.1, 0.5 \}$ and the regularization rate $\beta$ using the range of $\beta = \{ 0.1, 0.5, 1.0 \}$. For \quate and \bique, we tune the learning rate $\alpha$ using the range of $\alpha = \{ 0.1, 0.5, 1.0, 2.0, 5.0 \}$ and the regularization rate $\beta$ using the range of $\beta = \{ 0.1, 0.5, 1.0 \}$. For \transd, \anal, \quate, and \bique, validation is done every 50 epochs up to 500 epochs, and we select the best epoch based on the validation results. Table~\ref{tb:best_base} shows the best hyperparameters of these methods for the triplet prediction and the base-level link prediction. For the conditional link prediction problem, we combine the scores of the triplet prediction and the base-level link prediction as described in the main paper; we use the best hyperparameters of the triplet prediction and the base-level link prediction.

\begin{table}[h]
\small
\centering
\begin{tabular}{cccc}
\toprule
 & & Triplet Prediction & Base-Level Link Prediction \\
\midrule
\multirow{3}{*}{\ptranse} & \fbh & $\alpha=5\cdot10^{-6}, \gamma=0.5$ & $\alpha=10^{-6}, \gamma=0.5$ \\
 & \fbhe & $\alpha=5\cdot10^{-6}, \gamma=0.5$ & $\alpha=10^{-6}, \gamma=0.5$ \\
 & \dbhe & $\alpha=10^{-6}, \gamma=0.5$ & $\alpha=10^{-6}, \gamma=0.5$ \\
\midrule
\multirow{3}{*}{\rpje} & \fbh & $\alpha=10^{-6}, \gamma=1.0$ & $\alpha=10^{-6}, \gamma=1.0$ \\
 & \fbhe & $\alpha=10^{-6}, \gamma=1.0$ & $\alpha=10^{-6}, \gamma=1.0$ \\
 & \dbhe & $\alpha=10^{-6}, \gamma=1.0$ & $\alpha=10^{-6}, \gamma=1.0$ \\
\midrule
\multirow{3}{*}{\transd} & \fbh & $\alpha=0.5, \gamma=5.0$ & $\alpha=1.0, \gamma=5.0$ \\
 & \fbhe & $\alpha=0.1, \gamma=5.0$ & $\alpha=1.0, \gamma=5.0$ \\
 & \dbhe & $\alpha=0.5, \gamma=5.0$ & $\alpha=0.1, \gamma=5.0$ \\
\midrule
\multirow{3}{*}{\anal} & \fbh & $\alpha=0.01, \beta=0.01$ & $\alpha=0.01, \beta=0.1$ \\
 & \fbhe & $\alpha=0.01, \beta=0.01$ & $\alpha=0.01, \beta=0.1$ \\
 & \dbhe & $\alpha=0.05, \beta=0.1$ & $\alpha=0.01, \beta=0.1$ \\
\midrule
\multirow{3}{*}{\quate} & \fbh & $\alpha=1.0, \beta=0.05$ & $\alpha=0.1, \beta=0.05$ \\
 & \fbhe & $\alpha=1.0, \beta=0.01$ & $\alpha=0.1, \beta=0.05$ \\
 & \dbhe & $\alpha=0.5, \beta=0.05$ & $\alpha=0.5, \beta=0.1$ \\
\midrule
\multirow{3}{*}{\bique} & \fbh & $\alpha=0.1, \beta=0.01$ & $\alpha=0.5, \beta=0.05$ \\
 & \fbhe & $\alpha=0.1, \beta=0.01$ & $\alpha=0.5, \beta=0.05$ \\
 & \dbhe & $\alpha=0.1, \beta=0.05$ & $\alpha=0.5, \beta=0.1$ \\
\bottomrule
\end{tabular}
\caption{The best hyperparameters of \ptranse, \rpje, \transd, \anal, \quate, and \bique. $\alpha, \beta, \gamma$ indicate the learning rate, the regularization rate, and the margin, respectively.}
\label{tb:best_base}
\end{table}

\vspace{-0.2cm}
\subsection{Hyperparameters of \ours}
\vspace{-0.2cm}
Within \ours, we use the scoring function of \quate~\cite{quate} for \oursq and \bique~\cite{bique} for \oursb. On the validation set, we tune the learning rate $\alpha$, the regularization rate $\beta$, and the weights $\lambda_1$ and $\lambda_2$ in $L_{\text{\ours}}$. We use the search space of $\{ 0.2, 0.5, 1.0 \}$ for both $\lambda_1$ and $\lambda_2$. Validation is done every 50 epochs up to 500 epochs, and we select the best epoch based on the validation results. Table~\ref{tb:best_ours} shows the best hyperparameters of \ours on the validation sets.

\begin{table}[t]
\small
\centering
\begin{tabular}{cccccccccccccc}
\toprule
 & & \multicolumn{4}{c}{\fbh} & \multicolumn{4}{c}{\fbhe} & \multicolumn{4}{c}{\dbhe} \\
 & & $\alpha$ & $\beta$ & $\lambda_1$ & $\lambda_2$ & $\alpha$ & $\beta$ & $\lambda_1$ & $\lambda_2$ & $\alpha$ & $\beta$ & $\lambda_1$ & $\lambda_2$ \\
\midrule
\multirow{3}{*}{\oursq} & Triplet Prediction & 0.1 & 0.01 & 0.5 & 1.0 & 0.1 & 0.01 & 1.0 & 0.2 & 0.5 & 0.05 & 0.2 & 1.0 \\
 & Conditional Link Prediction & 0.1 & 0.01 & 1.0 & 0.2 & 0.1 & 0.01 & 1.0 & 0.2 & 0.5 & 0.01 & 1.0 & 0.2 \\
 & Base-Level Link Prediction & 0.1 & 0.05 & 1.0 & 0.2 & 0.1 & 0.05 & 0.5 & 0.2 & 0.5 & 0.1 & 0.5 & 0.2 \\
\midrule
\multirow{3}{*}{\oursb} & Triplet Prediction & 0.1 & 0.01 & 0.5 & 0.5 & 0.1 & 0.01 & 0.5 & 0.2 & 0.1 & 0.01 & 1.0 & 0.2 \\
 & Conditional Link Prediction & 0.1 & 0.01 & 1.0 & 0.2 & 0.1 & 0.01 & 1.0 & 0.2 & 0.5 & 0.01 & 1.0 & 0.2 \\
 & Base-Level Link Prediction & 0.1 & 0.05 & 0.5 & 0.2 & 0.1 & 0.05 & 0.5 & 0.5 & 0.5 & 0.1 & 0.2 & 1.0 \\
\bottomrule
\end{tabular}
\caption{The best hyperparameters of \ours on validation.}
\label{tb:best_ours}
\end{table}

\vspace{-0.2cm}
\section{Real-World Bi-Level Knowledge Graphs}
\vspace{-0.2cm}
We describe the details about how we create the three real-world bi-level knowledge graphs, \fbh, \fbhe, and \dbhe.

\vspace{-0.2cm}
\subsection{Base-Level Knowledge Graphs}
\vspace{-0.2cm}
We use \fbk~\cite{fb} as the base-level knowledge graphs for \fbh and \fbhe. \fbk is a standard benchmark knowledge graph dataset which is constructed by taking 401 most frequent relations and merging near-duplicate and inverse relations in \fbko~\cite{transe} from \fb~\cite{fbl}. 

We use a filtered version of \dbk~\cite{kblrn} which is constructed based on \dbp~\cite{dbp}. By following the strategy used in~\cite{nell}, we first remove the relations that are not in the form of \dbp URL, such as `http://www.w3.org/2000/01/rdf-schema\#seeAlso', since these types of relations do not have clear semantics. Then, we take the relations involved in more than 100 triplets and merge near-duplicate and inverse relations by following the strategy used in~\cite{fb}. For example, (Chevrolet, owningCompany, General\_Motors) and (Chevrolet, owner, General\_Motors) are merged.

\vspace{-0.2cm}
\subsection{Higher-Level Triplets}
\vspace{-0.2cm}
In Table~\ref{tb:hlt_full}, we show the higher-level relations and the corresponding examples of the higher-level triplets used to create \fbh, \fbhe, and \dbhe. While \fbhe contains all ten higher-level relations, \fbh contains only the first six higher-level relations. 

Among the higher-level relations in Table~\ref{tb:hlt_full}, WorksFor, SucceededBy, TransfersTo, and HigherThan in \fbhe and NextAlmaMater and TransfersTo in \dbhe require externally-sourced knowledge. For example, we crawled Wikipedia articles to find information about the (vice)presidents of the United States, the teams a player was playing for, and the alma mater of politicians. Also, to create $\langle T_i$, HigherThan, $T_j\rangle$ in \fbhe, we used the most recent rankings of Fortune 1000 and Times University Ranking. Table~\ref{tb:fbh_full} and Table~\ref{tb:dbh_full} show all types of higher-level triplets used to create \fbh, \fbhe, and \dbhe.

\begin{table*}[htbp]
\scriptsize
\centering
\setlength{\tabcolsep}{0.15em}
\begin{tabular}{ccll}
\toprule
& $\rhat$ & \multicolumn{1}{c}{$\langle T_i,\rhat,T_j \rangle$} & Description \\
\midrule
\multirow{20}{*}{\fbhe} & \multirow{2}{*}{PrerequisiteFor} & $T_i$: (BAFTA\_Award, Nominates, The\_King's\_Speech) & For The King's Speech to win BAFTA Award, BAFTA Award should nominate The \\
 & & $T_j$: (The\_King's\_Speech, Wins, BAFTA\_Award) & King's Speech. \\
\cdashline{2-4}
 & \multirow{2}{*}{EquivalentTo} & $T_i$: (Hillary\_Clinton, IsMarriedTo, Bill\_Clinton) & \multirow{2}{*}{The two triplets indicate the same information.} \\
 & & $T_j$: (Bill\_Clinton, IsMarriedTo, Hillary\_Clinton) & \\
\cdashline{2-4}
 & \multirow{2}{*}{ImpliesLocation} & $T_i$: (Sweden, CapitalIsLocatedIn, Stockholm) & \multirow{2}{*}{`The capital of Sweden is Stockholm' implies `Sweden contains Stockholm'.} \\
 & & $T_j$: (Sweden, Contains, Stockholm) & \\
\cdashline{2-4}
 & \multirow{2}{*}{ImpliesProfession} & $T_i$: (Liam\_Neeson, ActsIn, Love\_Actually) & \multirow{2}{*}{`Liam Neeson acts in Love Actually' implies `Liam Neeson is an actor'.} \\
 & & $T_j$: (Liam\_Neeson, IsA, Actor) & \\
\cdashline{2-4}
 & \multirow{2}{*}{ImpliesSports} & $T_i$: (Boston\_Red\_Socks, HasPosition, Infield) & \multirow{2}{*}{`Boston Red Socks has an infield position' implies `Boston Red Socks plays baseball'.} \\
 & & $T_j$: (Boston\_Red\_Socks, Plays, Baseball) & \\
\cdashline{2-4}
 & \multirow{2}{*}{NextEventPlace} & $T_i$: (1932\_Summer\_Olympics, IsHeldIn, Los\_Angeles) & Summer Olympics in 1932 and 1936 were held in Los Angeles and Berlin, respectively. \\
 & & $T_j$: (1936\_Summer\_Olympics, IsHeldIn, Berlin) & 1936 Summer Olympics is the next event of 1932 Summer Olympics.\\
\cdashline{2-4}
 & \multirow{2}{*}{WorksFor} & $T_i$: (Joe\_Biden, HoldsPosition, Vice\_President) & \multirow{2}{*}{Joe Biden was a vice president when Barack Obama was a president of the United States.} \\
 & & $T_j$: (Barack\_Obama, HoldsPosition, President) & \\
\cdashline{2-4}
 & \multirow{2}{*}{SucceededBy} & $T_i$: (George\_W.\_Bush, HoldsPosition, President) & \multirow{2}{*}{President Barack Obama succeeded President George W. Bush.} \\
 & & $T_j$: (Barack\_Obama, HoldsPosition, President) & \\
\cdashline{2-4}
 & \multirow{2}{*}{TransfersTo} & $T_i$: (David\_Beckham, PlaysFor, Real\_Madrid\_CF) & \multirow{2}{*}{David Beckham transferred from Real Madrid CF to LA Galaxy.} \\
 & & $T_j$: (David\_Beckham, PlaysFor, LA\_Galaxy) & \\
\cdashline{2-4}
 & \multirow{2}{*}{HigherThan} & $T_i$: (Walmart, IsRankedIn, Fortune\_500) & \multirow{2}{*}{Walmart is ranked higher than Bank of America in Fortune 500.} \\
 & & $T_j$: (Bank\_of\_America, IsRankedIn, Fortune\_500) & \\
\cdashline{1-4}
\multirow{16}{*}{\dbhe} & \multirow{2}{*}{EquivalentTo} & $T_i$: (David\_Beckham, IsMarriedTo, Victoria\_Beckham) & \multirow{2}{*}{The two triplets indicate the same information.} \\
 & & $T_j$: (Victoria\_Beckham, IsMarriedTo, David\_Beckham) & \\
\cdashline{2-4}
 & \multirow{2}{*}{ImpliesLanguage} & $T_i$: (Italy, HasOfficialLanguage, Italian\_Language) & `The official language of Italy is the Italian language' implies `The Italian language is \\
 & & $T_j$: (Italy, UsesLanguage, Italian\_Language) & used in Italy'. \\
\cdashline{2-4}
 & \multirow{2}{*}{ImpliesProfession} & $T_i$: (Psycho, IsDirectedBy, Alfred\_Hitchcock) & \multirow{2}{*}{`Psycho is directed by Alfred Hitchcock' implies `Alfred Hitchcock is a film producer'.} \\
 & & $T_j$: (Alfred\_Hitchcock, IsA, Film\_Producer) & \\
\cdashline{2-4}
 & \multirow{2}{*}{ImpliesLocation} & $T_i$: (Mariah\_Carey, LivesIn, New\_York\_City) & \multirow{2}{*}{`Mariah Carey lives in New York City' implies `Mariah Carey lives in New York'} \\
 & & $T_j$: (Mariah\_Carey, LivesIn, New\_York) & \\
\cdashline{2-4}
 & \multirow{2}{*}{ImpliesTimeZone} & $T_i$: (Czech\_Republic, TimeZone, Central\_European) & `Czech Republic is included in Central European Time Zone' implies `Prague is included \\
 & & $T_j$: (Prague, TimeZone, Central\_European) & in Central European Time Zone'. \\
\cdashline{2-4}
 & \multirow{2}{*}{ImpliesGenre} & $T_i$: (Pharrell\_Williams, Genre, Progressive\_Rock) & `Pharrell Williams is a progressive rock musician' implies `Pharrell Williams is a rock \\
 & & $T_j$: (Pharrell\_Williams, Genre, Rock\_Music) & musician'. \\
\cdashline{2-4}
 & \multirow{2}{*}{NextAlmaMater} & $T_i$: (Gerald\_Ford, StudiesIn, University\_of\_Michigan) & \multirow{2}{*}{Gerald Ford studied in University of Michigan. Then, he studied in Yale University.} \\
 & & $T_j$: (Gerald\_Ford, StudiesIn, Yale\_University) & \\
\cdashline{2-4}
 & \multirow{2}{*}{TransfersTo} & $T_i$: (Ronaldo, PlaysFor, FC\_Barcelona) & \multirow{2}{*}{Ronaldo transferred from FC Barcelona to Inter Millan.} \\
 & & $T_j$: (Ronaldo, PlaysFor, Inter\_Millan) & \\
\bottomrule
\end{tabular}
\caption{The higher-level relations and the corresponding examples of the higher-level triplets used to create \fbhe, and \dbhe.}
\label{tb:hlt_full}
\end{table*}

\begin{table*}[htbp]
\scriptsize
\centering
\setlength{\tabcolsep}{1em}
\begin{tabular}{cllc}
\toprule
 & & Example & Frequency \\
\midrule
\multirow{6}{*}{$\langle T_i$, PrerequisiteFor, $T_j \rangle$} & $T_i$: (Person A, DatesWith, Person B) & (Bruce\_Willis, DatesWith, Demi\_Moore) & \multirow{2}{*}{222} \\
 & $T_j$: (Person A, BreaksUpWith, Person B) & (Bruce\_Willis, BreaksUpWith, Demi\_Moore) & \\
\cdashline{2-4}
 & $T_i$: (Award A, Nominates, Work B) & (BAFTA\_Award, Nominates, The\_King's\_Speech) & \multirow{2}{*}{2,335} \\
 & $T_j$: (Work B, Wins, Award A) & (The\_King's\_Speech, Wins, BAFTA\_Award) & \\
\cdashline{2-4}
 & $T_i$: (Person A, HasNationality, Country B) & (Neymar, HasNationality, Brazil) & \multirow{2}{*}{109} \\
 & $T_j$: (Person A, PlaysFor, National Team of B) & (Neymar, PlaysFor, Brazil\_National\_Football\_Team) &  \\
\cdashline{1-4}
\multirow{8}{*}{$\langle T_i$, EquivalentTo, $T_j \rangle$} & $T_i$: (Person A, IsASiblingTo, Person B) & (Serena\_Williams, IsASiblingTo, Venus\_Williams) & \multirow{2}{*}{120} \\
 & $T_j$: (Person B, IsASiblingTo, Person A) & (Venus\_Williams, IsASiblingTo, Serena\_Williams) & \\
\cdashline{2-4}
 & $T_i$: (Person A, IsMarriedTo, Person B) & (Hillary\_Clinton, IsMarriedTo, Bill\_Clinton) & \multirow{2}{*}{352} \\
 & $T_j$: (Person B, IsMarriedTo, Person A) & (Bill\_Clinton, IsMarriedTo, Hillary\_Clinton) & \\
\cdashline{2-4}
 & $T_i$: (Person A, HasAFriendshipWith, Person B) & (Bob\_Dylan, HasAFriendshipWith, The\_Beatles) & \multirow{2}{*}{1,832} \\
 & $T_j$: (Person B, HasAFriendshipWith, Person A) & (The\_Beatles, HasAFriendshipWith, Bob\_Dylan) & \\
\cdashline{2-4}
 & $T_i$: (Person A, IsAPeerOf, Person B) & (Jimi\_Hendrix, IsAPeerOf, Eric\_Clapton) & \multirow{2}{*}{132} \\
 & $T_j$: (Person B, IsAPeerOf, Person A) & (Eric\_Clapton, IsAPeerOf, Jimi\_Hendrix) & \\
\cdashline{1-4}
\multirow{6}{*}{$\langle T_i$, ImpliesLocation, $T_j \rangle$} & $T_i$: (Location A, Contains, Location B) & (England, Contains, Warwickshire) & \multirow{2}{*}{2,415} \\
 & $T_j$: (Location containing A, Contains, Location in B) & (United\_Kingdom, Contains, Birmingham) & \\
\cdashline{2-4}
 & $T_i$: (Organization A, Headquarter, Location B) & (Kyoto\_University, Headquarter, Kyoto) & \multirow{2}{*}{820} \\
 & $T_j$: (Location B, Contains, Organization A) & (Kyoto, Contains, Kyoto\_University) & \\
\cdashline{2-4}
 & $T_i$: (Country A, CapitalIsLocatedIn, City B) & (Sweden, CapitalIsLocatedIn, Stockholm) & \multirow{2}{*}{83} \\
 & $T_j$: (Country A, Contains, City B) & (Sweden, Contains, Stockholm) & \\
\cdashline{1-4}
\multirow{10}{*}{$\langle T_i$, ImpliesProfession, $T_j \rangle$} & $T_i$: (Person A, IsA, Specialized Profession of B) & (Mariah\_Carey, IsA, Singer-songwriter) & \multirow{2}{*}{2,364} \\
 & $T_j$: (Person A, IsA, Profession B) & (Mariah\_Carey, IsA, Musician) & \\
\cdashline{2-4}
 & $T_i$: (Rock\&Roll Hall of Fame, Inducts, Person A) & (Rock\&Roll\_Hall\_of\_Fame, Inducts, Bob\_Dylan) & \multirow{2}{*}{66} \\
 & $T_j$: (Person A, IsA, Musician/Artist) & (Bob\_Dylan, IsA, Musician) & \\
\cdashline{2-4}
 & $T_i$: (Film A, IsWrittenBy, Person B) & (127\_Hours, IsWrittenBy, Danny\_Boyle) & \multirow{2}{*}{893} \\
 & $T_j$: (Person B, IsA, Writer/Film Producer) & (Danny\_Boyle, IsA, Film\_producer) & \\
\cdashline{2-4}
 & $T_i$: (Person A, ActsIn, Film B) & (Liam\_Neeson, ActsIn, Love\_Actually) & \multirow{2}{*}{10,864} \\
 & $T_j$: (Person A, IsA, Actor) & (Liam\_Neeson, IsA, Actor) & \\
\cdashline{2-4}
 & $T_i$: (Person A, HoldsPosition, Government Position B) & (Barack\_Obama, HoldsPosition, President) & \multirow{2}{*}{120} \\
 & $T_j$: (Person A, IsA, Politician) & (Barack\_Obama, IsA, Politician) & \\
\cdashline{1-4}
\multirow{6}{*}{$\langle T_i$, ImpliesSports, $T_j \rangle$} & $T_i$: (Team A, HasPosition, Position of B) & (Boston\_Red\_Socks, HasPosition, Infield) & \multirow{2}{*}{2,936} \\
 & $T_j$: (Team A, Plays, Sports B) & (Boston\_Red\_Socks, Plays, Baseball) & \\
\cdashline{2-4}
 & $T_i$: (League of A, Includes, Team B) & (National\_League, Includes, New\_York\_Mets) & \multirow{2}{*}{824} \\
 & $T_j$: (Team B, Plays, Sports A) & (New\_York\_Mets, Plays, Baseball) & \\
\cdashline{2-4}
 & $T_i$: (Team A, ParticipatesIn, Draft of B) & (Atlanta\_Braves, ParticipatesIn, MLB\_Draft) & \multirow{2}{*}{528} \\
 & $T_j$: (Team A, Plays, Sports B) & (Atlanta\_Braves, Plays, Baseball) & \\
\cdashline{1-4}
\multirow{2}{*}{$\langle T_i$, NextEventPlace, $T_j \rangle$} & $T_i$: (Event A, IsHeldIn, Location A) & (1932\_Summer\_Olympics, IsHeldIn, Los\_Angeles) & \multirow{2}{*}{47} \\
 & $T_j$: (Next Event of A, IsHeldIn, Location B) & (1936\_Summer\_Olympics, IsHeldIn, Berlin) & \\
\cdashline{1-4}
\multirow{2}{*}{$\langle T_i$, WorksFor, $T_j \rangle$} & $T_i$: (Person A, HoldsPosition, Vice President) & (Joe\_Biden, HoldsPosition, Vice\_President) & \multirow{2}{*}{13} \\
 & $T_j$: (Person B, HoldsPosition, President) & (Barack\_Obama, HoldsPosition, President) & \\
\cdashline{1-4}
\multirow{2}{*}{$\langle T_i$, SucceededBy, $T_j \rangle$} & $T_i$: (Person A, HoldsPosition, President/Vice President) & (George\_W.\_Bush, HoldsPosition, President) & \multirow{2}{*}{30} \\
 & $T_j$: (Person B, HoldsPosition, President/Vice President) & (Barack\_Obama, HoldsPosition, President) & \\
\cdashline{1-4}
\multirow{2}{*}{$\langle T_i$, TransfersTo, $T_j \rangle$} & $T_i$: (Person A, PlaysFor, Team B) & (David\_Beckham, PlaysFor, Real\_Madrid\_CF) & \multirow{2}{*}{377} \\
 & $T_j$: (Person A, PlaysFor, Team C) & (David\_Beckham, PlaysFor, LA\_Galaxy) & \\
\cdashline{1-4}
\multirow{2}{*}{$\langle T_i$, HigherThan, $T_j \rangle$} & $T_i$: (Item A, IsRankedIn, Ranking List C) & (Walmart, IsRankedIn, Fortune\_500) & \multirow{2}{*}{7,459} \\
 & $T_j$: (Item B, IsRankedIn, Ranking List C) & (Bank\_of\_America, IsRankedIn, Fortune\_500) & \\
\bottomrule
\end{tabular}
\caption{All types of higher-level triplets to create \fbh and \fbhe.}
\label{tb:fbh_full}
\end{table*}

\begin{table*}[htbp]
\scriptsize
\centering
\setlength{\tabcolsep}{0.45em}
\begin{tabular}{cllc}
\toprule
 & & Example & Frequency \\
\midrule
\multirow{6}{*}{$\langle T_i$, EquivalentTo, $T_j \rangle$} & $T_i$: (Person A, IsMarriedTo, Person B) & (Hillary\_Clinton, IsMarriedTo, Bill\_Clinton) & \multirow{2}{*}{314} \\
 & $T_j$: (Person B, IsMarriedTo, Person A) & (Bill\_Clinton, IsMarriedTo, Hillary\_Clinton) & \\
\cdashline{2-4}
 & $T_i$: (Location A, UsesLanguage, Language B) & (Brazil, UsesLanguage, Portuguese\_Language) & \multirow{2}{*}{120} \\
 & $T_j$: (Language B, IsSpokenIn, Location A) & (Portuguese\_Language, IsSpokenIn, Brazil) & \\
\cdashline{2-4}
 & $T_i$: (Person A, Influences, Person B) & (Baruch\_Spinoza, Influences, Immanuel\_Kant) & \multirow{2}{*}{394} \\
 & $T_j$: (Person B, IsInfluencedBy, Person A) & (Immanuel\_Kant, IsInfluencedBy, Baruch\_Spinoza) & \\
\cdashline{1-4}
\multirow{4}{*}{$\langle T_i$, ImpliesLanguage, $T_j \rangle$} & $T_i$: (Location A, HasOfficialLanguage, Language B) & (Italy, HasOfficialLanguage, Italian\_Language) & \multirow{2}{*}{196} \\
 & $T_j$: (Location A, UsesLanguage, Language B) & (Italy, UsesLanguage, Italian\_Language) & \\
\cdashline{2-4}
 & $T_i$: (Location A, UsesLanguage, Language B) & (United\_States, UsesLanguage, English\_Language) & \multirow{2}{*}{75} \\
 & $T_j$: (Location in A, UsesLanguage, Language B) & (California, UsesLanguage, English\_Language) & \\
\cdashline{1-4}
\multirow{12}{*}{$\langle T_i$, ImpliesProfession, $T_j \rangle$} & $T_i$: (Work A, MusicComposedBy, Person B) & (Forrest\_Gump, MusicComposedBy, Alan\_Silvestri) & \multirow{2}{*}{553} \\
 & $T_j$: (Person B, IsA, Musician/Composer) & (Alan\_Silvestri, IsA, Composer) & \\
\cdashline{2-4}
 & $T_i$: (Work A, Starring, Person B) & (Love\_Actually, Starring, Liam\_Neeson) & \multirow{2}{*}{737} \\
 & $T_j$: (Person B, IsA, Actor) & (Liam\_Neeson, IsA, Actor) & \\
\cdashline{2-4}
 & $T_i$: (Work A, CinematographyBy, Person B) & (Jurassic\_Park, CinematographyBy, Dean\_Cundey) & \multirow{2}{*}{299} \\
 & $T_j$: (Person B, IsA, Cinematographer) & (Dean\_Cundey, IsA, Cinematographer) & \\
\cdashline{2-4}
 & $T_i$: (Work A, IsDirectedBy, Person B) & (Psycho, IsDirectedBy, Alfred\_Hitchcock) & \multirow{2}{*}{295} \\
 & $T_j$: (Person B, IsA, Film\_Director/Television\_Director) & (Alfred\_Hitchcock, IsA, Film\_Director) & \\
\cdashline{2-4}
 & $T_i$: (Work A, IsProducedBy, Person B) & (King\_Kong, IsProducedBy, Merian\_C.\_Cooper) & \multirow{2}{*}{354} \\
 & $T_j$: (Person B, IsA, Film\_Producer/Television\_Producer) & (Merian\_C.\_Cooper, IsA, Film\_Producer) & \\
\cdashline{2-4}
 & $T_i$: (Person A, AssociatesWithRecordLabel, Record B) & (Bo\_Diddley, AssociatesWithRecordLabel, Atlantic\_Records) & \multirow{2}{*}{155} \\
 & $T_j$: (Person A, IsA, Record\_Producer) & (Bo\_Diddley, IsA, Record\_Producer) & \\
\cdashline{1-4}
\multirow{6}{*}{$\langle T_i$, ImpliesLocation, $T_j \rangle$} & $T_i$: (Location A, IsPartOf, Location B) & (Ann\_Arbor, IsPartOf, Washtenaw\_County\_Michigan) & \multirow{2}{*}{1,174} \\
 & $T_j$: (Location in A, IsPartOf, Location Containing B) & (Ann\_Arbor, IsPartOf, Michigan) & \\
\cdashline{2-4}
 & $T_i$: (Organization A, IsLocatedIn, Location B) & (Adobe\_Systems, IsLocatedIn, San\_Jose\_California) & \multirow{2}{*}{250} \\
 & $T_j$: (Organization A, IsLocatedIn, Location Containing B) & (Adobe\_Systems, IsLocatedIn, California) & \\
\cdashline{2-4}
 & $T_i$: (Person A, LivesIn, Location B) & (Mariah\_Carey, LivesIn, New\_York\_City) & \multirow{2}{*}{213} \\
 & $T_j$: (Person A, LivesIn, Location Containing B) & (Mariah\_Carey, LivesIn, New\_York) & \\
\cdashline{1-4}
\multirow{2}{*}{$\langle T_i$, ImpliesTimeZone, $T_j \rangle$} & $T_i$: (Location A, TimeZone, Time Zone B) & (Czech\_Republic, TimeZone, Central\_European\_Time) & \multirow{2}{*}{409} \\
 & $T_j$: (Location in A, TimeZone, Time Zone B) & (Prague, TimeZone, Central\_European\_Time) & \\
\cdashline{1-4}
\multirow{2}{*}{$\langle T_i$, ImpliesGenre, $T_j \rangle$} & $T_i$: (Musician A, Genre, Genre B) & (Pharrell\_Williams, Genre, Progressive\_Rock) & \multirow{2}{*}{767} \\
 & $T_j$: (Musician A, Genre, Parent Genre of B) & (Pharrell\_Williams, Genre, Rock\_Music) & \\
\cdashline{1-4}
\multirow{2}{*}{$\langle T_i$, NextAlmaMater, $T_j \rangle$} & $T_i$: (Person A, StudiesIn, Institution B) & (Gerald\_Ford, StudiesIn, University\_of\_Michigan) & \multirow{2}{*}{112} \\
 & $T_j$: (Person A, StudiesIn, Institution C) & (Gerald\_Ford, StudiesIn, Yale\_University) & \\
\cdashline{1-4}
\multirow{2}{*}{$\langle T_i$, TransfersTo, $T_j \rangle$} & $T_i$: (Person A, PlaysFor, Team B) & (Ronaldo, PlaysFor, FC\_Barcelona) & \multirow{2}{*}{300} \\
 & $T_j$: (Person A, PlaysFor, Team C) & (Ronaldo, PlaysFor, Inter\_Millan) & \\
\bottomrule
\end{tabular}
\caption{All types of higher-level triplets to create \dbhe.}
\label{tb:dbh_full}
\end{table*}

\clearpage
\section*{Additional Experimental Results}
\label{sec:new}
We provide additional experimental results using a different implementation of \ours. The implementation of BiVE is based on OpenKE~\cite{openke}. Three loss functions, $L_{\text{base}}$, $L_{\text{high}}$, and $L_{\text{aug}}$, are implemented based on the Softplus loss provided in OpenKE, which can be formulated as follows:
\begin{equation}
L=\frac{1}{|\sE|}\sum_{(h,r,t)\in\sE}g(-f(\vh,\vr,\vt))+\frac{1}{|\sE'|}\sum_{(h',r',t')\in\sE'}g(f(\vh',\vr',\vt'))
\label{eq:loss}
\end{equation}

We recently implemented a different version of the Softplus loss that averages the loss incurred by the positive and negative triplets at once. The new implementation of the Softplus loss can be formulated as follows:
\begin{displaymath}
L_\text{new}=\frac{1}{|\sE|+|\sE'|}\left[\sum_{(h,r,t)\in\sE}g(-f(\vh,\vr,\vt))+\sum_{(h',r',t')\in\sE'}g(f(\vh',\vr',\vt'))\right]
\end{displaymath}

Using the new implementation of the loss in \ours, we provide the experimental results of triplet prediction and conditional link prediction in Table~\ref{tb:tp_new} and Table~\ref{tb:clp_new}. Since \anal, \quate, and \bique also use the Softplus loss implemented in OpenKE, the results of these methods are also changed. We see that the new implementation of the Softplus loss improves the performance of \ours. Also, Table~\ref{tb:blp_new} shows the performance of the typical base-level link prediction. In all experiments, the conclusion remains the same; our \ours models significantly outperform baseline methods for the triplet prediction and conditional link prediction tasks while achieving comparable results to the baseline methods for the base-level link prediction task. Table~\ref{tb:ab_new} shows the results of the ablation study of \ours with the new implementation.

Note that the original implementation of \ours is also correct; the loss term defined in (\ref{eq:loss}) aims to make the scores of the positive triplets higher than those of the negative triplets. Both implementations of \ours can be found at \url{https://github.com/bdi-lab/BiVE}.

\begin{table*}[h]
\small
\centering
\setlength{\tabcolsep}{0.22em}
\begin{tabular}{cccccccccc}
\toprule
 & \multicolumn{3}{c}{\fbh} & \multicolumn{3}{c}{\fbhe} & \multicolumn{3}{c}{\dbhe} \\
 & MR ($\downarrow$) & MRR ($\uparrow$) & Hit@10 ($\uparrow$) & MR ($\downarrow$) & MRR ($\uparrow$) & Hit@10 ($\uparrow$) & MR ($\downarrow$) & MRR ($\uparrow$) & Hit@10 ($\uparrow$) \\
\midrule
\aser & 74541.7{\scriptsize$\pm$0.0} & 0.011{\scriptsize$\pm$0.000} & 0.015{\scriptsize$\pm$0.000} & 57916.0{\scriptsize$\pm$0.0} & 0.050{\scriptsize$\pm$0.000} & 0.070{\scriptsize$\pm$0.000} & 18157.6{\scriptsize$\pm$0.0} & 0.042{\scriptsize$\pm$0.000} & 0.075{\scriptsize$\pm$0.000} \\
\mnv & 109055.1{\scriptsize$\pm$98.5} & 0.093{\scriptsize$\pm$0.002} & 0.113{\scriptsize$\pm$0.002} & 85571.5{\scriptsize$\pm$768.3} & 0.220{\scriptsize$\pm$0.008} & 0.300{\scriptsize$\pm$0.005} & 20764.3{\scriptsize$\pm$72.3} & 0.177{\scriptsize$\pm$0.005} & 0.221{\scriptsize$\pm$0.004} \\
\mhop & 108731.7{\scriptsize$\pm$43.2} & 0.105{\scriptsize$\pm$0.001} & 0.117{\scriptsize$\pm$0.000} & 83643.8{\scriptsize$\pm$33.2} & 0.255{\scriptsize$\pm$0.012} & 0.311{\scriptsize$\pm$0.003} & 20505.8{\scriptsize$\pm$9.3} & 0.191{\scriptsize$\pm$0.001} & 0.230{\scriptsize$\pm$0.002} \\
\nelp & 115016.6{\scriptsize$\pm$0.0} & 0.070{\scriptsize$\pm$0.000} & 0.073{\scriptsize$\pm$0.000} & 90000.4{\scriptsize$\pm$0.0} & 0.238{\scriptsize$\pm$0.000} & 0.274{\scriptsize$\pm$0.000} & 21130.5{\scriptsize$\pm$0.0} & 0.170{\scriptsize$\pm$0.000} & 0.209{\scriptsize$\pm$0.000} \\
\drum & 115016.6{\scriptsize$\pm$0.0} & 0.069{\scriptsize$\pm$0.001} & 0.073{\scriptsize$\pm$0.000} & 90000.3{\scriptsize$\pm$0.0} & 0.261{\scriptsize$\pm$0.000} & 0.274{\scriptsize$\pm$0.000} & 21130.5{\scriptsize$\pm$0.0} & 0.166{\scriptsize$\pm$0.001} & 0.209{\scriptsize$\pm$0.000} \\
\anyb & 108079.6{\scriptsize$\pm$0.0} & 0.096{\scriptsize$\pm$0.000} & 0.108{\scriptsize$\pm$0.000} & 83136.8{\scriptsize$\pm$5.3} & 0.191{\scriptsize$\pm$0.001} & 0.252{\scriptsize$\pm$0.001} & 20530.8{\scriptsize$\pm$0.0} & 0.177{\scriptsize$\pm$0.000} & 0.214{\scriptsize$\pm$0.000} \\
\ptranse & 111024.3{\scriptsize$\pm$855.0} & 0.069{\scriptsize$\pm$0.000} & 0.071{\scriptsize$\pm$0.000} & 86793.2{\scriptsize$\pm$961.0} & 0.249{\scriptsize$\pm$0.001} & 0.274{\scriptsize$\pm$0.000} & 18888.7{\scriptsize$\pm$457.3} & 0.158{\scriptsize$\pm$0.001} & 0.195{\scriptsize$\pm$0.002} \\
\rpje & 113082.0{\scriptsize$\pm$945.2} & 0.070{\scriptsize$\pm$0.000} & 0.072{\scriptsize$\pm$0.000} & 89173.1{\scriptsize$\pm$797.3} & 0.267{\scriptsize$\pm$0.000} & 0.274{\scriptsize$\pm$0.000} & 20290.4{\scriptsize$\pm$417.2} & 0.166{\scriptsize$\pm$0.001} & 0.206{\scriptsize$\pm$0.002} \\
\transd & 74277.3{\scriptsize$\pm$2907.8} & 0.052{\scriptsize$\pm$0.001} & 0.104{\scriptsize$\pm$0.002} & 52159.4{\scriptsize$\pm$758.9} & 0.238{\scriptsize$\pm$0.002} & 0.280{\scriptsize$\pm$0.003} & 16698.1{\scriptsize$\pm$370.2} & 0.116{\scriptsize$\pm$0.004} & 0.189{\scriptsize$\pm$0.009} \\
\anal & 152635.3{\scriptsize$\pm$554.7} & 0.100{\scriptsize$\pm$0.001} & 0.110{\scriptsize$\pm$0.001} & 118023.1{\scriptsize$\pm$337.9} & 0.284{\scriptsize$\pm$0.003} & 0.310{\scriptsize$\pm$0.001} & 23512.7{\scriptsize$\pm$3265.1} & 0.160{\scriptsize$\pm$0.004} & 0.199{\scriptsize$\pm$0.008} \\
\quate & 109954.7{\scriptsize$\pm$2068.8} & 0.104{\scriptsize$\pm$0.000} & 0.114{\scriptsize$\pm$0.001} & 85021.3{\scriptsize$\pm$1402.8} & 0.251{\scriptsize$\pm$0.013} & 0.282{\scriptsize$\pm$0.005} & 27548.3{\scriptsize$\pm$304.0} & 0.163{\scriptsize$\pm$0.001} & 0.191{\scriptsize$\pm$0.002} \\
\bique & 79802.8{\scriptsize$\pm$528.9} & 0.104{\scriptsize$\pm$0.000} & 0.115{\scriptsize$\pm$0.000} & 59997.8{\scriptsize$\pm$519.0} & 0.293{\scriptsize$\pm$0.002} & 0.319{\scriptsize$\pm$0.000} & 18259.8{\scriptsize$\pm$231.0} & 0.160{\scriptsize$\pm$0.001} & 0.194{\scriptsize$\pm$0.001} \\
\oursq & \textbf{5.6{\scriptsize$\pm$0.9}} & \textbf{0.876{\scriptsize$\pm$0.003}} & \textbf{0.938{\scriptsize$\pm$0.003}} & \textbf{10.7{\scriptsize$\pm$9.3}} & \textbf{0.728{\scriptsize$\pm$0.008}} & \textbf{0.882{\scriptsize$\pm$0.013}} & \textbf{4.3{\scriptsize$\pm$0.4}} & \textbf{0.634{\scriptsize$\pm$0.008}} & \textbf{0.923{\scriptsize$\pm$0.005}} \\
\oursb & \underline{7.9{\scriptsize$\pm$1.2}} & \underline{0.862{\scriptsize$\pm$0.008}} & \underline{0.931{\scriptsize$\pm$0.005}} & \underline{17.7{\scriptsize$\pm$23.0}} & \underline{0.708{\scriptsize$\pm$0.008}} & \underline{0.863{\scriptsize$\pm$0.012}} & \underline{11.6{\scriptsize$\pm$5.7}} & \underline{0.629{\scriptsize$\pm$0.018}} & \underline{0.867{\scriptsize$\pm$0.021}} \\
\bottomrule
\end{tabular}
\caption{Results of Triplet Prediction with New Implementation. The best scores are boldfaced and the second best scores are underlined. Our models, \oursq and \oursb, significantly outperform all other baseline methods in terms of all metrics on all datasets.}
\label{tb:tp_new}
\end{table*}

\clearpage
\begin{table*}[h]
\small
\centering
\setlength{\tabcolsep}{0.42em}
\begin{tabular}{cccccccccc}
\toprule
 & \multicolumn{3}{c}{\fbh} & \multicolumn{3}{c}{\fbhe} & \multicolumn{3}{c}{\dbhe} \\
 & MR ($\downarrow$) & MRR ($\uparrow$) & Hit@10 ($\uparrow$) & MR ($\downarrow$) & MRR ($\uparrow$) & Hit@10 ($\uparrow$) & MR ($\downarrow$) & MRR ($\uparrow$) & Hit@10 ($\uparrow$) \\
\midrule
\aser & 1183.9{\scriptsize$\pm$0.0} & 0.251{\scriptsize$\pm$0.000} & 0.316{\scriptsize$\pm$0.000} & 970.7{\scriptsize$\pm$0.0} & 0.289{\scriptsize$\pm$0.000} & 0.382{\scriptsize$\pm$0.000} & 1893.5{\scriptsize$\pm$0.0} & 0.225{\scriptsize$\pm$0.000} & 0.348{\scriptsize$\pm$0.000} \\
\mnv & 3817.8{\scriptsize$\pm$58.9} & 0.328{\scriptsize$\pm$0.013} & 0.415{\scriptsize$\pm$0.009} & 3018.5{\scriptsize$\pm$45.8} & 0.407{\scriptsize$\pm$0.013} & 0.492{\scriptsize$\pm$0.014} & 2934.1{\scriptsize$\pm$32.2} & 0.362{\scriptsize$\pm$0.007} & 0.433{\scriptsize$\pm$0.014} \\
\mhop & 1878.2{\scriptsize$\pm$12.0} & 0.421{\scriptsize$\pm$0.003} & 0.578{\scriptsize$\pm$0.003} & 1447.3{\scriptsize$\pm$11.9} & 0.443{\scriptsize$\pm$0.002} & 0.615{\scriptsize$\pm$0.002} & 1012.3{\scriptsize$\pm$28.5} & 0.442{\scriptsize$\pm$0.007} & 0.652{\scriptsize$\pm$0.008} \\
\nelp & 185.9{\scriptsize$\pm$1.3} & 0.433{\scriptsize$\pm$0.002} & 0.648{\scriptsize$\pm$0.004} & 146.2{\scriptsize$\pm$1.0} & 0.466{\scriptsize$\pm$0.002} & 0.716{\scriptsize$\pm$0.007} & 32.2{\scriptsize$\pm$1.9} & 0.517{\scriptsize$\pm$0.006} & 0.756{\scriptsize$\pm$0.004} \\
\drum & 262.7{\scriptsize$\pm$13.3} & 0.394{\scriptsize$\pm$0.002} & 0.555{\scriptsize$\pm$0.003} & 207.6{\scriptsize$\pm$10.0} & 0.413{\scriptsize$\pm$0.010} & 0.620{\scriptsize$\pm$0.018} & 49.0{\scriptsize$\pm$3.9} & 0.470{\scriptsize$\pm$0.010} & 0.732{\scriptsize$\pm$0.012} \\
\anyb & 228.5{\scriptsize$\pm$11.8} & 0.380{\scriptsize$\pm$0.004} & 0.563{\scriptsize$\pm$0.013} & 166.0{\scriptsize$\pm$7.9} & 0.418{\scriptsize$\pm$0.002} & 0.607{\scriptsize$\pm$0.008} & 81.7{\scriptsize$\pm$4.0} & 0.403{\scriptsize$\pm$0.002} & 0.594{\scriptsize$\pm$0.004} \\
\ptranse & 214.8{\scriptsize$\pm$0.7} & 0.440{\scriptsize$\pm$0.001} & 0.686{\scriptsize$\pm$0.002} & 167.0{\scriptsize$\pm$1.8} & 0.516{\scriptsize$\pm$0.001} & 0.752{\scriptsize$\pm$0.001} & 19.3{\scriptsize$\pm$0.2} & 0.505{\scriptsize$\pm$0.004} & 0.780{\scriptsize$\pm$0.001} \\
\rpje & 212.5{\scriptsize$\pm$0.1} & 0.440{\scriptsize$\pm$0.001} & 0.686{\scriptsize$\pm$0.001} & 159.0{\scriptsize$\pm$0.0} & 0.528{\scriptsize$\pm$0.001} & 0.753{\scriptsize$\pm$0.001} & 19.3{\scriptsize$\pm$0.1} & 0.504{\scriptsize$\pm$0.004} & 0.779{\scriptsize$\pm$0.002} \\
\transd & 190.1{\scriptsize$\pm$18.0} & 0.300{\scriptsize$\pm$0.003} & 0.496{\scriptsize$\pm$0.005} & 165.6{\scriptsize$\pm$8.0} & 0.363{\scriptsize$\pm$0.003} & 0.529{\scriptsize$\pm$0.006} & 35.5{\scriptsize$\pm$1.0} & 0.436{\scriptsize$\pm$0.006} & 0.708{\scriptsize$\pm$0.005} \\
\anal & 130.6{\scriptsize$\pm$11.6} & 0.331{\scriptsize$\pm$0.014} & 0.486{\scriptsize$\pm$0.033} & 122.4{\scriptsize$\pm$21.8} & 0.335{\scriptsize$\pm$0.036} & 0.501{\scriptsize$\pm$0.055} & 67.3{\scriptsize$\pm$83.9} & 0.391{\scriptsize$\pm$0.129} & 0.600{\scriptsize$\pm$0.224} \\
\quate & 124.4{\scriptsize$\pm$1.5} & 0.399{\scriptsize$\pm$0.004} & 0.572{\scriptsize$\pm$0.009} & 94.6{\scriptsize$\pm$1.4} & 0.419{\scriptsize$\pm$0.003} & 0.598{\scriptsize$\pm$0.008} & 31.0{\scriptsize$\pm$1.8} & 0.432{\scriptsize$\pm$0.013} & 0.700{\scriptsize$\pm$0.019} \\
\bique & 107.5{\scriptsize$\pm$1.3} & 0.414{\scriptsize$\pm$0.001} & 0.640{\scriptsize$\pm$0.003} & 84.9{\scriptsize$\pm$0.7} & 0.444{\scriptsize$\pm$0.003} & 0.670{\scriptsize$\pm$0.003} & 20.1{\scriptsize$\pm$3.0} & 0.493{\scriptsize$\pm$0.003} & 0.775{\scriptsize$\pm$0.003} \\
\oursq & \textbf{2.2{\scriptsize$\pm$0.1}} & \textbf{0.913{\scriptsize$\pm$0.005}} & \textbf{0.982{\scriptsize$\pm$0.001}} & \textbf{3.8{\scriptsize$\pm$0.2}} & \textbf{0.838{\scriptsize$\pm$0.003}} & \textbf{0.929{\scriptsize$\pm$0.003}} & \textbf{2.3{\scriptsize$\pm$0.2}} & \textbf{0.860{\scriptsize$\pm$0.003}} & \textbf{0.978{\scriptsize$\pm$0.003}} \\
\oursb & \underline{2.7{\scriptsize$\pm$0.3}} & \underline{0.907{\scriptsize$\pm$0.004}} & \underline{0.978{\scriptsize$\pm$0.002}} & \underline{4.3{\scriptsize$\pm$0.3}} & \underline{0.833{\scriptsize$\pm$0.002}} & \underline{0.928{\scriptsize$\pm$0.003}} & \underline{3.3{\scriptsize$\pm$0.4}} & \underline{0.845{\scriptsize$\pm$0.012}} & \underline{0.960{\scriptsize$\pm$0.005}} \\
\bottomrule
\end{tabular}
\caption{Results of Conditional Link Prediction with New Implementation. The best scores are boldfaced and the second best scores are underlined. Our models, \oursq and \oursb, significantly outperform all other baseline methods in terms of all metrics on all datasets.}
\label{tb:clp_new}
\end{table*}

\begin{multicols}{2}
\balance
\begin{table}[H]
\scriptsize
\centering
\setlength{\tabcolsep}{0.76em}
\begin{tabular}{ccccc}
\toprule
 & \multicolumn{2}{c}{\fbhe} & \multicolumn{2}{c}{\dbhe} \\
 & MR ($\downarrow$) & Hit@10 ($\uparrow$) & MR ($\downarrow$) & Hit@10 ($\uparrow$) \\
\midrule
\aser & 1489.3{\scriptsize$\pm$0.0} & 0.323{\scriptsize$\pm$0.000} & 2218.8{\scriptsize$\pm$0.0} & 0.197{\scriptsize$\pm$0.000}\\
\mnv & 3828.4{\scriptsize$\pm$56.9} & 0.339{\scriptsize$\pm$0.003} & 3530.7{\scriptsize$\pm$50.1} & 0.297{\scriptsize$\pm$0.006}\\
\mhop & 2284.0{\scriptsize$\pm$9.5} & 0.500{\scriptsize$\pm$0.001} & 2489.4{\scriptsize$\pm$15.3} & 0.404{\scriptsize$\pm$0.004}\\
\nelp & 1942.5{\scriptsize$\pm$0.5} & 0.486{\scriptsize$\pm$0.001} & 2904.8{\scriptsize$\pm$0.6} & 0.357{\scriptsize$\pm$0.001}\\
\drum & 1945.6{\scriptsize$\pm$0.8} & 0.490{\scriptsize$\pm$0.002} & 2904.7{\scriptsize$\pm$0.7} & 0.359{\scriptsize$\pm$0.001}\\
\anyb & 342.0{\scriptsize$\pm$4.6} & 0.526{\scriptsize$\pm$0.002} & 879.1{\scriptsize$\pm$5.7} & 0.364{\scriptsize$\pm$0.003}\\
\ptranse & 2077.6{\scriptsize$\pm$10.3} & 0.333{\scriptsize$\pm$0.000} & 3346.0{\scriptsize$\pm$20.0} & 0.277{\scriptsize$\pm$0.002}\\
\rpje & 1754.6{\scriptsize$\pm$7.5} & 0.368{\scriptsize$\pm$0.001} & 2991.7{\scriptsize$\pm$28.1} & 0.341{\scriptsize$\pm$0.000}\\
\transd & 166.3{\scriptsize$\pm$1.3} & 0.527{\scriptsize$\pm$0.001} & \textbf{429.0{\scriptsize$\pm$7.5}} & 0.423{\scriptsize$\pm$0.001} \\
\anal & 244.2{\scriptsize$\pm$7.7} & 0.516{\scriptsize$\pm$0.003} & 1049.5{\scriptsize$\pm$47.2} & 0.332{\scriptsize$\pm$0.006} \\
\quate & 144.1{\scriptsize$\pm$2.6} & 0.594{\scriptsize$\pm$0.001} & 549.1{\scriptsize$\pm$11.3} & 0.451{\scriptsize$\pm$0.002} \\
\bique & 140.4{\scriptsize$\pm$1.7} & 0.591{\scriptsize$\pm$0.001} & \underline{505.2{\scriptsize$\pm$5.6}} & \textbf{0.458{\scriptsize$\pm$0.001}} \\
\oursq & \underline{127.6{\scriptsize$\pm$2.6}} & \underline{0.596{\scriptsize$\pm$0.002}} & 552.3{\scriptsize$\pm$10.8} & \underline{0.453{\scriptsize$\pm$0.001}} \\
\oursb & \textbf{124.8{\scriptsize$\pm$1.6}} & \textbf{0.598{\scriptsize$\pm$0.001}} & 524.1{\scriptsize$\pm$8.6} & 0.448{\scriptsize$\pm$0.002} \\
\bottomrule
\end{tabular}
\caption{Results of Base-Level Link Prediction with New Implementation.}
\label{tb:blp_new}
\end{table}

\begin{table}[H]
\small
\centering
\begin{tabular}{ccccc}
\toprule
 & & \fbh & \fbhe & \dbhe \\
\midrule
\multirow{2}{*}{TP} & $L_{\text{base}} + L_{\text{high}}$ & 5.1 & 11.9 & 4.1 \\
& $L_{\text{base}} + L_{\text{high}}+ L_{\text{aug}}$ & 5.6 & 10.7 & 4.3 \\
\cdashline{1-5}
\multirow{2}{*}{CLP} & $L_{\text{base}} + L_{\text{high}}$ & 2.7 & 4.3 & 2.2 \\
& $L_{\text{base}} + L_{\text{high}}+ L_{\text{aug}}$ & 2.2 & 3.8 & 2.3 \\
\cdashline{1-5}
\multirow{4}{*}{BLP} & $L_{\text{base}}$ & 144.1 & 144.1 & 549.1 \\
& $L_{\text{base}} + L_{\text{high}}$ & 141.4 & 143.3 & 563.5 \\
& $L_{\text{base}} + L_{\text{aug}}$ & 127.9 & 127.7 & 541.6 \\
& $L_{\text{base}} + L_{\text{high}}+ L_{\text{aug}}$ & 126.0 & 127.6 & 552.3 \\
\bottomrule
\end{tabular}
\caption{Ablation study of \ours with different combinations of the loss terms (with the new implementation of \ours). The average MR scores on triplet prediction (TP), conditional link prediction (CLP), and the base-level link prediction (BLP).}
\label{tb:ab_new}
\end{table}

\end{multicols}
\end{document}